%% file: main.tex
\documentclass{article}

\usepackage[preprint]{neurips_2026}
\usepackage[utf8]{inputenc} 
\usepackage[T1]{fontenc}    
\usepackage{hyperref}       
\usepackage{url}            
\usepackage{booktabs}       
\usepackage{amsfonts}       
\usepackage{nicefrac}     
\usepackage{microtype}     
\usepackage{xcolor}         
\usepackage{multirow}
\usepackage{array}
\usepackage{graphicx}
\usepackage[most]{tcolorbox}
\usepackage{listings}
\usepackage{xcolor}
\usepackage{graphicx}
\usepackage{xcolor}
\AddToHook{shipout/background}{}
\pdfpageattr{/Group <</S /Transparency /I true /CS /DeviceRGB>>}
\usepackage[utf8]{inputenc} 
\usepackage{graphicx}     
\usepackage{subcaption}    
\usepackage{float}        
\usepackage{algorithm}
\usepackage{algorithmic}
\usepackage{xcolor}
\usepackage{amsmath}
\usepackage{amssymb}
\usepackage{enumitem}
\usepackage{wrapfig}
\usepackage{hyperref}
\usepackage{marvosym}
\usepackage{pifont}
\definecolor{takeawaycolor}{RGB}{40,40,40}
\newcommand{\takeaway}[2]{%
\vspace{3pt}
\noindent
{\large\color{takeawaycolor}\ding{228}}\;
\textbf{Takeaway #1.}~#2
\vspace{3pt}
}

\colorlet{punct}{red!60!black}
\definecolor{background}{HTML}{F8F8F8}
\definecolor{delim}{RGB}{20,105,176}
\colorlet{numb}{magenta!60!black}

\newcommand{\placeholderblock}[1]{%
  \par\medskip
  \noindent
  \begin{center}
  \tcbox[
    colback=yellow!12,
    colframe=orange!80!black,
    boxrule=0.4pt,
    arc=1pt,
    left=4pt,
    right=4pt,
    top=2pt,
    bottom=2pt
  ]{\texttt{\textless{}#1\textgreater{}}}
  \end{center}
  \medskip
}

\lstdefinelanguage{json}{
    basicstyle=\normalfont\ttfamily\small,
    numbers=left,
    numberstyle=\scriptsize\color{gray},
    stepnumber=1,
    numbersep=8pt,
    showstringspaces=false,
    breaklines=true,
    frame=lines,
    backgroundcolor=\color{background},
    literate=
     *{0}{{{\color{numb}0}}}{1}
      {1}{{{\color{numb}1}}}{1}
      {2}{{{\color{numb}2}}}{1}
      {3}{{{\color{numb}3}}}{1}
      {4}{{{\color{numb}4}}}{1}
      {5}{{{\color{numb}5}}}{1}
      {6}{{{\color{numb}6}}}{1}
      {7}{{{\color{numb}7}}}{1}
      {8}{{{\color{numb}8}}}{1}
      {9}{{{\color{numb}9}}}{1}
      {:}{{{\color{punct}{:}}}}{1}
      {,}{{{\color{punct}{,}}}}{1}
      {\{}{{{\color{delim}{\{}}}}{1}
      {\}}{{{\color{delim}{\}}}}}{1}
      {[}{{{\color{delim}{[}}}}{1}
      {]}{{{\color{delim}{]}}}}{1},
}

\title{ToolAnchor: Anchoring Counterfactual Context to Boost Agentic Tool-use Capability}

\newcommand\blfootnote[1]{%
  \begingroup
  \renewcommand\thefootnote{}\footnotetext{#1}%
  \endgroup
}

\author{%
  \textbf{Weiting Liu}$^{1,2}$, \
  \textbf{Jieyi Bi}$^{3}$, \
  \textbf{Wanqi Zhou}$^{2}$, \
  \textbf{Jianfeng Feng}$^{1}$, \\
  \textbf{Yining Ma}$^{4,\mbox{\scriptsize\Letter}}$, \
  \textbf{Ai Han}$^{2,\mbox{\scriptsize\Letter}}$, \
  \textbf{Wenlian Lu}$^{1,\dagger}$ \\[8pt]
  \small
  $^{1}$Fudan University \quad
  $^{2}$JD.COM \quad
  $^{3}$Nanyang Technological University \quad
  $^{4}$MIT
}

\begin{document}

\maketitle

\blfootnote{$^{\dagger}$Project Leader. \quad \Letter: Corresponding authors: Yining Ma (\texttt{yiningma@mit.edu}), Ai Han (\texttt{hanai5@jd.com}).}

\begin{abstract}

Tool-augmented large language model agents excel at long-horizon tasks, yet they are typically post-trained on fixed toolsets. When tasks demand new tools, these agents struggle to incorporate them effectively, and retraining from scratch is often impractical.
We identify the core obstacle in such toolset expansion problem as \textit{behavioral inertia}: the tendency of agents to fall back on familiar tools and established reasoning patterns despite having access to new ones. We demonstrate that injecting counterfactual anchor contexts at critical decision points can break this inertia, recovering failed trajectories by eliciting suppressed agent capabilities. To scale this insight, we propose ToolAnchor, a framework that uses teacher models to \textit{hypothesize} these counterfactual contexts, \textit{verifies} them via student rollouts, and \textit{internalizes} the successful interventions through agentic post-training. Extensive evaluations across general AI assistant (GAIA), textual search (BrowseComp), and visual search (VDR-Bench) tasks demonstrate that ToolAnchor consistently exhibits competitive performance under expanded toolsets. Our work bridges the gap between static post-training and dynamic adaptation, charting a new path for scalable agentic reinforcement learning.

\end{abstract}

\section{Introduction}
\label{sec:intro}

\input{Sections/introduction}

\section{Related work}
\label{sec:relatedwork}

\input{Sections/related_works}

\section{Preliminaries}
\label{sec:preliminary_analysis}

\input{Sections/Preliminaries}

\section{Methodology}
\label{sec:methodology}

\input{Sections/methodology}

\section{Experiments}
\label{sec:exp}

\input{Sections/experiments}

\section{Conclusion and limitation}
\label{sec:conclusion}
\input{Sections/conclusion}

\clearpage
\newpage

{
\bibliographystyle{unsrt}
\bibliography{main}
}

\clearpage
\newpage
\appendix

\input{appendix}

\end{document}

%% file: Sections/introduction.tex
Recent advances in tool-augmented agents have significantly broadened the real-world utility of Large Language Models (LLMs)~\cite{singh2025agentic,jiang2025verltool,li2025reinforcement,fan2026exploring,li2026reasoning,fang2026proximity}. By integrating external tools into the reasoning loop, LLM-based agents have achieved strong performance across a wide spectrum of downstream tasks, spanning text-based question answering~\cite{wu2025webwalker,wei2025browsecomp,zhou2025browsecomp,mialon2023gaia}, visual question answering~\cite{yang2025magic,huang2026vision,zhang2026visbrowse}, long-form text report generation~\cite{li2025webweaver,du2025deepresearch}, and domain-grounded tool use~\cite{yao2024tau,barres2025tau}. Central to this progress are the ReAct paradigm~\cite{yao2022react}, which interleaves reasoning with tool invocation, and agentic reinforcement learning (agentic RL) methods~\cite{shao2024deepseekmath,yu2025dapo,zhang2025landscape}, which further refine tool-use capabilities via post-training. However, these successes largely assume a fixed, predefined toolset available throughout training. In practice, task requirements evolve continuously, often necessitating new tools such as web search for enhanced retrieval~\cite{jin2025search,li2025search}, python interpreters for scientific computation~\cite{feng2025retool,xue2025simpletir}, file parsers for document understanding~\cite{team2025tongyi}, and visual tools for multimodal perception~\cite{huang2026vision,fan2026exploring}.

A natural question then arises: \textbf{\textit{how can an existing agent efficiently adapt to an expanded toolset?}} The most straightforward approach is to directly apply agentic RL, allowing the agent to discover effective reasoning strategies and action policies via exploration. However, this strategy faces a fundamental difficulty. Since the agent's policy has already converged during prior agentic RL training, it tends to fall back on familiar behavioral patterns, underutilizing newly introduced tools and repeating suboptimal reasoning strategies. This \textit{behavioral inertia} leads to sparse rewards and inefficient exploration, as the agent rarely ventures beyond its established routines. An alternative is to rely on the tool-use generalization capability of the agent~\cite{zou2025autotool,du2025generalizable}, hoping that it can adapt to new tools without additional training. Yet, as our preliminary analysis in Section~\ref{sec:preliminary_analysis} reveals, generalization alone is insufficient. To better understand this bottleneck, we conduct a motivation study on failed student trajectories. A teacher model first identifies a decision round where an alternative local thought--action pair could plausibly redirect the subsequent rollout; we call this decision point a \textit{counterfactual anchor round} (Section~\ref{sec:counterfactual_anchor}). The teacher then proposes this alternative using only the history before that round, yielding a teacher-revised context from which the student resumes generation. Compared with resuming from the original context, these teacher-revised contexts improve task success in both textual and visual search settings. This suggests that failures arise not only from limited tool-use capability, but also from missing contextual anchors that redirect reasoning and tool selection at key decision points.

\begin{figure}[tbp!]
    \centering
    \includegraphics[width=0.9\linewidth]{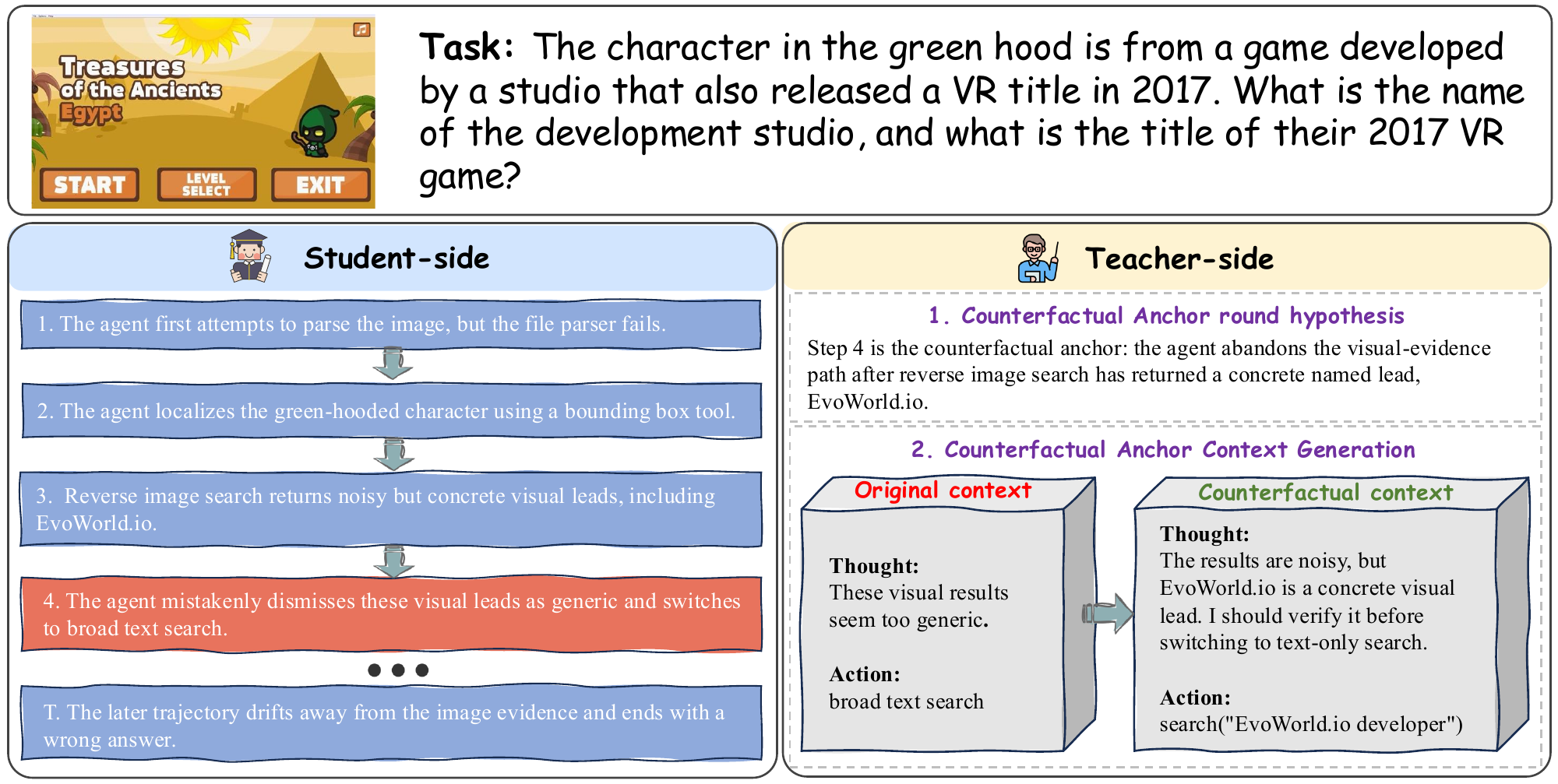}
    \caption{A case study of counterfactual anchor context generation}
    \label{fig:toolanchor_teaser_figure}
\end{figure}

In this work, we propose \textbf{ToolAnchor}, a framework that bootstraps new tool-use capabilities through verified counterfactual context anchoring. ToolAnchor operates in three stages. In the \textit{Counterfactual Anchor hypothesis} stage, diverse teacher models audit the student agent's failed rollout trajectories and hypothesize anchor rounds where the agent's decision deviated from a promising path. While the primary focus is on rounds where the agent failed to invoke a newly introduced tool, the teacher also considers suboptimal uses of existing tools and flawed reasoning steps, as these often compound the difficulty of adopting new tools effectively. For each identified anchor round, the teacher hypothesizes alternative reasoning and actions that would plausibly redirect the trajectory toward task completion. In the \textit{Counterfactual Anchor Verification} stage, the student agent resumes generation from these teacher-anchored contexts under agentic rollout. Crucially, we retain for training only those anchor contexts that satisfy both verification criteria, namely task-level success and anchor-level usefulness, rather than accepting teacher hypotheses uncritically. In the \textit{Counterfactual Anchor Internalization} stage, the verified contexts are used for agentic supervised fine-tuning (agentic SFT), allowing the agent to absorb the anchored behaviors into its own policy. This is followed by agentic RL, which realigns the internalized anchor behaviors with the agent's previously learned agentic policy and supports robust, autonomous use of the expanded toolset.

Our contributions are summarized as follows:
(\textbf{1}) We formalize the practical challenge of expanding a post-trained agent's toolset. Through a motivation study, we identify \textit{behavioral inertia} as the core bottleneck and reveal that counterfactual contextual anchoring at critical decision points can effectively redirect failed trajectories.
(\textbf{2}) We propose ToolAnchor, a scalable, three-stage framework for capability expansion. It automates the discovery of counterfactual anchor rounds via diverse teacher models, strictly verifies their effectiveness through student rollouts, and internalizes these validated behaviors via agentic SFT and RL.
(\textbf{3}) Extensive evaluations across general AI assistance (GAIA), textual search (BrowseComp), and visual search (VDR-Bench) demonstrate that ToolAnchor seamlessly integrates new visual tools. Furthermore, it simultaneously boosts the base agent's reasoning and existing tool mastery, offering a robust paradigm for continuous agentic evolution.

%% file: Sections/related_works.tex
\paragraph{Tool-augmented deep research agents.}

Deep research agents tackle open-ended question answering through long-horizon tool use: iteratively planning, invoking external tools, verifying evidence, performing intermediate reasoning, and synthesizing final answers~\cite{li2025reinforcement}. This paradigm naturally encompasses general AI assistance, textual search, and visual search. Textual search remains the most extensively explored direction, where agents retrieve external evidence to answer complex queries. For instance, Search-R1~\cite{jin2025search}, Search-o1~\cite{li2025search}, and WebThinker~\cite{li2025webthinker} directly integrate web search into reasoning, while subsequent work advances this via agentic SFT and RL~\cite{wu2025webdancer}, systematic uncertainty reduction~\cite{li2025websailor}, scalable data synthesis~\cite{qiao2025webresearcher}, dual-environment RL frameworks~\cite{li2025websailor}, and continual pre-training~\cite{su2025scaling}. Building on these techniques, the Tongyi DeepResearch Agent~\cite{team2025tongyi} achieves competitive performance, and OpenSeeker~\cite{du2026openseeker} provides high-quality training datasets. Visual search extends this paradigm to queries requiring visual evidence, demanding agents to not only retrieve web pages but also understand and ground visual content in external knowledge. Benchmarks like VDR-Bench~\cite{zeng2026vision} and VisBrowse~\cite{zhang2026visbrowse} evaluate this capability, alongside VLM-driven agents such as T3-Agent~\cite{gao2024multi} and Vision-DeepResearch~\cite{huang2026vision} which explore multimodal tool integration. However, existing visual search approaches predominantly build agents from scratch with jointly designed toolsets, rather than extending well-trained text-oriented agents. This leaves a practical gap: given the immense cost of training capable deep research agents, incrementally augmenting them with visual tools is far more practical than starting over. We address this gap as a toolset expansion problem.

\paragraph{Agentic reinforcement learning.}

    Reinforcement learning has emerged as a powerful post-training paradigm for LLM-based agents~\cite{zhang2025landscape,yang2026learning,zhang2026heterogeneous}. GRPO~\cite{shao2024deepseekmath} introduced group relative policy optimization for mathematical reasoning, and DAPO~\cite{yu2025dapo} further enhanced its performance through a series of techniques. Although these methods can be applied to agentic RL settings, multi-turn tool-integrated reasoning introduces distinct challenges, such as distributional shift caused by tool feedback that deviates from the LLM's training data. To mitigate these issues, several methods have been proposed: SimpleTIR~\cite{xue2025simpletir} filters trajectories containing void turns, GiGPO computes both episode-level and step-level relative advantages, and ARPO~\cite{dong2025agentic} and AEPO~\cite{dong2025agentic_aepo} use entropy-based adaptive beaming strategies to execute partial rollouts at high-entropy tokens. Despite these advances, existing agentic RL methods share a common assumption: the toolset is fixed throughout training. When new tools are introduced to a post-trained agent, its policy has already converged around the original toolsets, and standard agentic RL exploration mechanisms may be insufficient to overcome this \textit{behavioral inertia}. Our work targets this setting through a teacher-guided anchoring mechanism that is then validated by student rollout. This design contrasts with standard knowledge distillation~\cite{hinton2015distilling} and on-policy imitation methods such as DAgger~\cite{ross2011reduction}, which treat teacher-generated reasoning and actions as ground-truth supervision. Our framework treats teacher guidance as a hypothesis to be verified through student rollouts rather than as a direct supervised target.

%% file: Sections/preliminaries.tex
\subsection{Tool-use agent setup}
\label{sec:agent_setup}

Following Tongyi DeepResearch Agent~\cite{team2025tongyi}, we consider a post-trained tool-use agent equipped with an original toolset $U_{\mathrm{old}}$. In our experiments, $U_{\mathrm{old}}$ consists of web search, web browsing, a Python interpreter, scholar search, and a file parser. To study toolset expansion, we introduce a set of new visual tools $U_{\mathrm{new}}$, including an image descriptor, a bounding-box extractor, and crop-and-image search. The resulting expanded toolset is $U = U_{\mathrm{old}} \cup U_{\mathrm{new}}$. Detailed tool descriptions are provided in Appendix~\ref{appendix:tool_description}. We follow the ReAct framework, which consists of thought, action, and observation in each round. At each round, the agent first generates a thought and an action, and then receives an observation from the environment. This process forms a trajectory, denoted as $\mathcal{H}_T$, which is a sequence of thought–action–observation triplets: $\mathcal{H}_T = (\tau_0, a_0, o_0, \ldots, \tau_i, a_i, o_i, \ldots, \tau_T, a_T)$, where $a_T$ represents the final answer to the given question. For each non-terminal round \(t<T\), the agent samples \((\tau_t,a_t)\sim \pi_\theta(\cdot\mid \mathcal H_{t-1})\), after which the environment executes \(a_t\) and returns \(o_t\). The terminal action \(a_T\) corresponds to the final answer, and no observation \(o_T\) is returned.

\subsection{Counterfactual anchor contexts}
\label{sec:counterfactual_anchor}

While existing diagnostics (e.g., AgentErrorBench~\cite{zhang2025agentic}) focus on categorizing root-cause errors in failed trajectories, we move beyond passive diagnosis to active intervention. Specifically, we investigate whether replacing a single local decision under a shared history can redirect the rollout and recover a failed task. Aligning with recent advances in agentic context engineering~\cite{zhang2025agentic,dou2026cl}, this motivates our formulation of the counterfactual anchor context. Formally, given a failed trajectory $\mathcal{H}_T$ and a non-terminal round $t<T$, the original local decision is the thought-action pair $(\tau_t,a_t)$ generated under history $\mathcal{H}_{t-1}$. A counterfactual alternative retains $\mathcal{H}_{t-1}$ unchanged but replaces this decision with a new pair $(\tau'_t,a'_t)$, conditioned solely on the available information in $\mathcal{H}_{t-1}$. The environment then executes $a'_t$ and returns the corresponding observation $o'_t$, yielding the counterfactual anchor context $c_t = (\mathcal{H}_{t-1}, \tau'_t, a'_t, o'_t)$. Crucially, $c_t$ isolates the divergent branch: it captures the precise local context arising from the alternative decision $(\tau'_t,a'_t)$ without altering prior history. We term $c_t$ an \emph{anchor} because it serves as a revised contextual reset point, rather than a full trajectory, from which the student agent resumes generation

\subsection{Preliminary analysis}
\label{sec:preliminary_analysis}

    To investigate tool-use adaptation, we benchmark the text-oriented Tongyi DeepResearch Agent, extended with three new visual tools (image descriptor, bbox extractor, crop-and-image search), against two general-purpose models (Gemini-2.5-Pro, Qwen3-30B-A3B-Instruct) on 100 random instances from VDR-Bench. As detailed in Table 1, the tool-call frequencies reveal a stark contrast: while general-purpose agents seamlessly adopt the visual tools, Tongyi remains heavily reliant on its original toolset. This discrepancy highlights a pronounced behavioral inertia in post-trained agents.

\input{Tables/motivation_table1}

    \takeaway{1}{
    The tool-call statistics support the \textit{behavioral inertia} hypothesis: under an expanded toolset, the post-trained student agent still relies disproportionately on the original tools and uses the newly introduced visual tools less frequently than general-purpose agents.
    }

    We further conduct motivation experiments on the textual search dataset OpenSeeker~\cite{du2026openseeker} and the visual search dataset VDR-Bench~\cite{huang2026vision}. Specifically, we randomly sample 100 failed trajectories of the Tongyi DeepResearch Agent from OpenSeeker and VDR-Bench, respectively. For each failed trajectory, we use teacher models (Gemini-2.5-Pro and GPT-5) to hypothesize the counterfactual anchor round and generate an alternative thought-action pair at that round (see Appendix~\ref{appendix:prompt_anchor_hypothesis} for the prompt). We then have the Tongyi DeepResearch Agent resume generation from the teacher-revised context. Table~\ref{tab:guess_anchor} compares the task success rate when the agent continues from the original context versus the teacher-revised context. The results show that resuming generation from the teacher-revised context improves task success accuracy.

    \input{Tables/motivation_table2}

    \takeaway{2}{
    Teacher-hypothesized \textit{counterfactual anchor contexts} can partially recover failed trajectories: compared with resuming from the original context, resuming from teacher-revised contexts yields higher task success in both textual and visual search settings.
    }

    These two findings motivate ToolAnchor, which systematically hypothesizes critical anchor rounds, verifies the counterfactual contexts, and internalizes the verified alternatives into the student agent.

%% file: Tables/motivation_table1.tex
\begin{table*}[htbp!]
    \centering
    \caption{Average number of tool calls per sampled instance from VDR-Bench}
    \label{tab:tool_call_statistics}
    \small
    \resizebox{\textwidth}{!}{
    \begin{tabular}{clccc}
        \toprule
        Toolset & Tool & Gemini-2.5-Pro & Qwen3-30B-A3B-Instruct & Tongyi-DeepResearch-30B-A3B \\
        \midrule
        \multirow{5}{*}{$U_{\mathrm{old}}$}
        & Web Search & 7.50 & 5.14 & 24.18  \\
        & Python Interpreter & 0.95 & 0.00 & 0.11  \\
        & Scholar Search & 0.16 & 0.47 & 0.00  \\
        & Web Browse & 4.42 & 2.72 & 9.62  \\
        & File Parser & 0.95 & 0.05 & 0.87  \\
        \midrule
        \multirow{3}{*}{$U_{\mathrm{new}}$}
        & Image Descriptor & 1.88 & 2.37 & 1.84  \\
        & Bbox Extractor & 1.43 & 1.68 & 0.37  \\
        & Crop-and-Image Search & 1.69 & 1.02 & 0.78  \\
        \bottomrule
    \end{tabular}
    }
\end{table*}

%% file: Tables/motivation_table2.tex
\begin{table*}[htbp!]
    \centering
    \caption{Effect of the counterfactual anchor context on task success rate}
    \label{tab:guess_anchor}
    \small
    \resizebox{\textwidth}{!}{
    \begin{tabular}{lcc}
        \toprule
        Anchor & OpenSeeker (Textual Search) & VDR-Bench (Visual Search) \\
        \midrule
        Student original context & 4.0\% &  11.0\% \\
        Teacher hypothesized context (Gemini-2.5-Pro) & 9.0\% &  23.0\% \\
        Teacher hypothesized context (GPT-5) & 6.0\% &  17.0\% \\
        \bottomrule
    \end{tabular}
    }
\end{table*}

%% file: Sections/methodology.tex
In this section, we present \textbf{ToolAnchor}, a framework for adapting a post-trained tool-use agent to an expanded toolset through counterfactual anchor contexts. ToolAnchor consists of three stages: counterfactual anchor context hypothesis, counterfactual anchor context verification, and counterfactual anchor context internalization.

\begin{figure}[htbp!]
    \centering
    \includegraphics[width=0.95\linewidth]{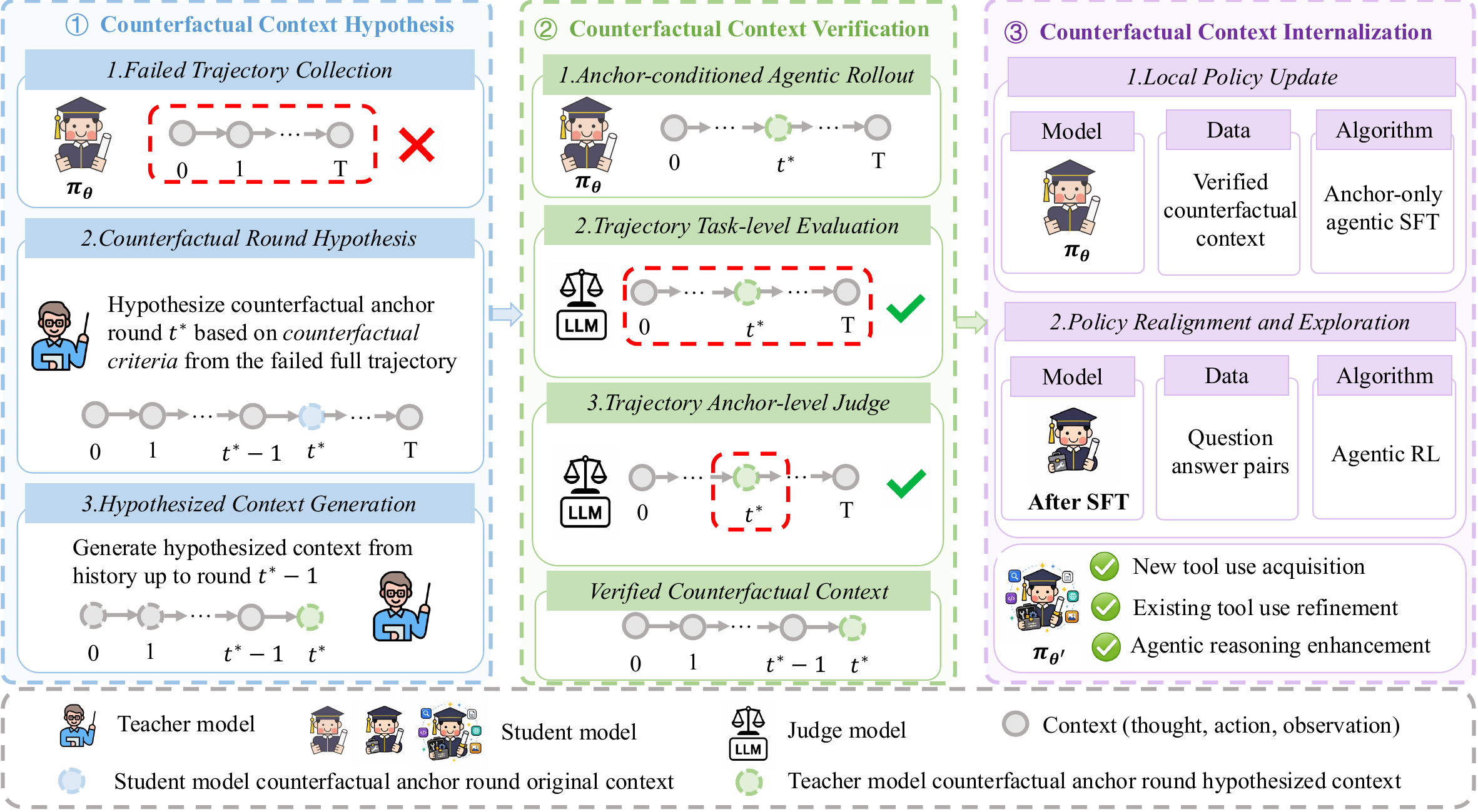}
    \caption{Overview of the ToolAnchor framework}
    \label{fig:pipeline}
\end{figure}

\subsection{Problem formulation}
\label{sec:problem}

    Building on the setup in Section~\ref{sec:agent_setup}, we study toolset expansion for a post-trained tool-use agent $\pi_\theta$. The agent has been trained with the original toolset $U_{\mathrm{old}}$ and is then exposed to the expanded toolset $U = U_{\mathrm{old}} \cup U_{\mathrm{new}}$. The goal is to learn an updated policy $\pi_{\theta'}$ that improves task success under $U$ while preserving the agent's existing capability on tasks solvable with $U_{\mathrm{old}}$.

    Our preliminary analysis in Section~\ref{sec:preliminary_analysis} reveals two key findings. First, the post-trained agent exhibits \textit{behavioral inertia}, disproportionately relying on $U_{\mathrm{old}}$ even when $U_{\mathrm{new}}$ is available (Table~\ref{tab:tool_call_statistics}). Second, failed trajectories can often be recovered when the agent resumes generation from a teacher-proposed counterfactual anchor context at a critical decision round (Table~\ref{tab:guess_anchor}). These findings suggest that the bottleneck is not only missing tool-use capability, but also the lack of effective contextual anchors that redirect the agent away from its established behavioral patterns. This motivates ToolAnchor, which hypothesizes, verifies, and internalizes such contexts, as illustrated in Figure~\ref{fig:pipeline}.

\subsection{Counterfactual anchor context hypothesis}
\label{sec:guess}

    Given a failed trajectory $\mathcal{H}_T$ produced by the base policy $\pi_\theta$ under the expanded toolset $U$, this stage uses teacher models to form a counterfactual anchor hypothesis. The hypothesis consists of a candidate anchor round $t^*<T$ and an alternative thought-action pair $(\tau'_{t^*}, a'_{t^*})$, which together define a revised context that may redirect the subsequent student rollout toward a successful trajectory.

    \paragraph{Failed trajectory collection.}

        We first roll out the base policy $\pi_\theta$ on the training datasets under the expanded toolset $U$, collecting both successful and failed trajectories. We define the failed trajectory set as $\mathcal{D}_{\mathrm{fail}} = \{\mathcal{H}_T \mid \mathrm{Success}(\mathcal{H}_T)=0\}$, which serves as the input for the teacher-guided counterfactual anchor hypothesis.

    \paragraph{Counterfactual round hypothesis.}

        We define an anchor round hypothesis as a candidate round at which an alternative local decision may induce a different subsequent rollout and potentially change the task outcome. Forming this hypothesis requires assessing how earlier decisions, observations, and subsequent reasoning steps interact across the trajectory, rather than relying on a fixed heuristic such as choosing the first or last suspicious step. We instantiate this step with a prompt-guided teacher audit: given the full failed trajectory, a teacher model sequentially reviews each round's thought $\tau_t$, action $a_t$, and observation $o_t$, and selects a candidate anchor round $t^*$ where a revised context is likely to redirect the subsequent rollout. The prompt used in our implementation is provided in Appendix~\ref{appendix:prompt_anchor_hypothesis}. In our implementation, the teacher is instructed to use the following criteria:

        \begin{enumerate}[label=\textit{Criterion \arabic*:}, leftmargin=*]
            \item \textit{Downstream impact.} Prefer a round whose alternative local decision is likely to influence multiple subsequent steps in the trajectory.
        
            \item \textit{Path redirection.} Prefer a round that acts as a branching point, where a different local decision may redirect the subsequent rollout toward a different reasoning or tool-use path.
        
            \item \textit{Strategic leverage.} Prefer a round involving high-level agentic decisions, such as reasoning direction, tool selection, query formulation, evidence interpretation, information integration, or planning.
        
            \item \textit{Counterfactual potential.} Prefer a round where replacing the original local decision under the same history appears most likely to improve the final task outcome.
        \end{enumerate}

        These criteria are intended to avoid trivial anchor-selection heuristics, such as always choosing the earliest suspicious step or the final decision step, both of which may fail to identify the round with the greatest potential to redirect the subsequent rollout.

    \paragraph{Hypothesized context generation.}

        Once the anchor round $t^*$ is hypothesized, the teacher model generates the alternative thought-action pair in a separate step using only the history available before that round. Specifically, the teacher is given only $\mathcal{H}_{t^*-1}$, without access to subsequent rounds or the ground-truth answer. The teacher then proposes an alternative thought-action pair $(\tau'_{t^*}, a'_{t^*}) \sim M(\cdot \mid \mathcal{H}_{t^*-1})$, where $M$ denotes the teacher model.

        This separation is important: the anchor-round hypothesis can use the full failed trajectory for diagnosis, whereas the alternative thought-action pair must be generated under the same information constraint faced by the agent at round \(t^*\). This prevents the alternative context from encoding future observations or the ground-truth answer, making the resulting supervision compatible with the student's inference-time information state. Although new-tool adoption is a central motivation, the teacher is not restricted to rounds involving $U_{\mathrm{new}}$. It may also select rounds involving ineffective use of $U_{\mathrm{old}}$, or reasoning steps that make the subsequent rollout less effective. The environment then executes $a'_{t^*}$ and returns the corresponding observation $o'_{t^*}$. This produces the candidate anchor context $c^* = (\mathcal{H}_{t^*-1}, \tau'_{t^*}, a'_{t^*}, o'_{t^*})$.

    \paragraph{Teacher diversity.}

        A single teacher model may favor particular tool-use patterns or anchor-round choices when auditing failed trajectories. To improve coverage, we use a set of heterogeneous teacher models $\mathcal{M}=\{M_j\}_{j=1}^J$, each of which independently forms counterfactual anchor hypotheses for the same failed trajectory. Different teachers may select different anchor rounds or propose different alternative thought-action pairs, yielding a diverse set of candidate anchor contexts $\mathcal{D}_{\mathrm{anchor}}=\{c_i^*\}_{i=1}^m$. Each $c_i^*$ is a teacher-hypothesized candidate counterfactual context and will be verified by student rollout in Section~\ref{sec:verify}.

\subsection{Counterfactual anchor context verification}
\label{sec:verify}

    Teacher-generated anchor contexts are hypotheses rather than ground-truth demonstrations. A teacher may select an ineffective anchor round, propose a plausible but unhelpful alternative decision, or generate an observation that does not support the student's subsequent reasoning. To ensure data quality without manual annotation or heuristic reward shaping, we introduce a student-side verification stage. This stage retains an anchor context only when the student successfully solves the task after resuming from it and a separate judge model, with access to the completed trajectory, verifies that the anchor context was materially useful for the final success.

    \paragraph{Anchor-conditioned agentic rollout and dual-level verification.}
        
        For each candidate anchor context $c^* \in \mathcal{D}_{\mathrm{anchor}}$, we let the base policy $\pi_\theta$ resume generation from $c^*$ under the standard agentic rollout procedure, producing a complete trajectory $\mathcal{H}'_T = \mathrm{Rollout}(\pi_\theta, c^*)$. We retain $c^*$ only if it passes two verification checks. First, the final answer in $\mathcal{H}'_T$ must be judged correct by the task-level evaluator, denoted as $J_{\mathrm{task}}(\mathcal{H}'_T) \in \{0,1\}$. Second, an anchor-level judge, denoted as $J_{\mathrm{anchor}}(c^*, \mathcal{H}'_T) \in \{0,1\}$, verifies from a global view that the proposed anchor context was materially useful in enabling the successful outcome. Formally, the verified effective-context dataset is
        
        \begin{equation}
            \mathcal{D}_{\mathrm{eff}} =
            \{(c^*, \mathcal{H}'_T) \mid
            \mathcal{H}'_T = \mathrm{Rollout}(\pi_\theta, c^*),
            J_{\mathrm{task}}(\mathcal{H}'_T) = 1,
            J_{\mathrm{anchor}}(c^*, \mathcal{H}'_T) = 1
            \}.
        \end{equation}

    \paragraph{Why verification matters.}

        Verification turns teacher hypotheses into training signals. The task-level success check filters out anchors from which the student cannot solve the task, and the anchor-level judge removes cases where success is not materially supported by the proposed anchor. Since the continuation after $c^*$ is generated by the student policy $\pi_\theta$, the retained data avoid full teacher-trajectory imitation and remain closer to the student's rollout distribution.

\subsection{Counterfactual anchor context internalization}
\label{sec:internalize}

    The final stage internalizes verified counterfactual anchor contexts into the agent's policy through a two-step pipeline, combining local policy internalization and policy alignment with exploration.

    \paragraph{Local policy update (anchor-only agentic SFT).}

        We perform anchor-only agentic SFT on the verified effective contexts, training the agent to reproduce the teacher-proposed thought-action pair at the verified counterfactual anchor round. For each verified counterfactual context $(c^*, \mathcal{H}'_T) \in \mathcal{D}_{\mathrm{eff}}$, where $c^*=(\mathcal{H}_{t^*-1}, \tau'_{t^*}, a'_{t^*}, o'_{t^*})$, the SFT loss is computed only on the alternative thought-action pair $(\tau'_{t^*}, a'_{t^*})$, conditioned on the original history $\mathcal{H}_{t^*-1}$:

        \begin{equation}
            \mathcal{L}_{\mathrm{SFT}}
            =
            -\mathbb{E}_{(c^*, \mathcal{H}'_T) \in \mathcal{D}_{\mathrm{eff}}}
            \left[
            \log \pi_\theta(\tau'_{t^*}, a'_{t^*} \mid \mathcal{H}_{t^*-1})
            \right].
        \end{equation}

        This objective directly injects the verified anchor behaviors at high-leverage decision points, providing a precise behavioral guidance. While optimizing solely on $\mathcal{D}_\text{eff}$ could risk localized overfitting, this anchor-only SFT serves as a targeted policy initialization. The subsequent post-SFT agentic RL stage then realigns these local updates with the agent's prior broad capabilities, effectively mitigating catastrophic forgetting and preserving performance on tasks solvable with $U_\text{old}$.

    \paragraph{Policy realignment and exploration (post-SFT agentic RL).}

        Following agentic SFT, we apply agentic RL (GRPO) on the expanded toolset $U$, using a binary task-level reward (1 for correct answers, 0 otherwise). This stage serves to integrate the internalized anchor behaviors from SFT with the agent’s prior RL policy, consolidating these local updates into a policy that is both aligned with prior knowledge and capable of effective inference under the expanded toolset. Simultaneously, it enables the agent to explore additional tool-use strategies beyond those explicitly provided by the teachers. Collectively, agentic SFT and agentic RL ensure that local counterfactual behaviors are internalized, properly aligned with the prior policy, and generalized, allowing the agent to leverage both newly introduced tools and previously learned capabilities.

%% file: Sections/experiments.tex
\subsection{Experimental setup}
\label{sec:setup}
 
    \paragraph{Student and teacher models.}

        We use the Tongyi DeepResearch Agent~\cite{team2025tongyi} (Qwen3-30B-A3B) as our base model $\pi_\theta$, which is trained via agentic RL on a fixed toolset $U_{\text{old}}$ consisting of web search, web browse, a Python interpreter, scholar search, and a file parser. We extend it with three visual tools $U_{\text{new}}$: an image descriptor, a bbox extractor, and crop-and-image search. We choose the Tongyi DeepResearch Agent not only because its system prompts, toolsets, and model weights are open-source (allowing for training and customization), but also because it demonstrates strong performance on general AI assistant and textual search tasks. For the counterfactual anchor hypothesis, we employ Gemini-2.5-Pro and GPT-5 as diverse teacher models to identify anchor rounds and generate alternative reasoning and actions.

    \paragraph{Training datasets and benchmarks.}

        We use two training datasets that span both textual and visual search tasks: (1) OpenSeeker-1.5k, which contains 1,500 instances sampled from OpenSeeker-v1-Data~\cite{du2026openseeker}; and (2) VDR-1.5k~\cite{huang2026vision}, a visual search dataset that also contains 1,500 samples. We evaluate our method on three benchmarks corresponding to distinct tasks: GAIA~\cite{mialon2023gaia} (General AI assistant, 165 samples), BrowseComp~\cite{wei2025browsecomp} (Textual search, 200 samples subset following~\cite{du2026openseeker}), and VDR-Bench~\cite{zeng2026vision} (Visual search, test-mini split with 500 samples, following~\cite{huang2026vision}).

    \paragraph{Evaluation metrics and baselines.}

        Following prior work~\cite{huang2026vision,du2026openseeker}, we adopt an LLM-as-a-judge approach for evaluation, using Qwen3-235B-A22B as the judge model. Specifically, we use the official prompts provided by each benchmark to assess the correctness of the agent’s final answer against the ground truth. Our primary evaluation metric is task success rate under Pass@1, where each query is evaluated with a single agent rollout. All benchmark results are reported with Pass@1. Pass@3 is used only during counterfactual anchor verification, where a candidate context passes task-level verification if at least one of three student rollouts solves the task. Please refer to Appendix~\ref{appendix:prompt_evaluation} for the corresponding evaluation prompts. We compare ToolAnchor against both proprietary and open-source deep research agents, including GPT-5~\cite{singh2025openai}, Gemini-2.5-Pro~\cite{comanici2025gemini}, Claude-4-Sonnet, OpenAI DeepResearch, Qwen3-VL-30B-A3B-Thinking~\cite{bai2025qwen3}, Qwen3-VL-30B-A3B-Instruct, OpenSeeker~\cite{du2026openseeker}, Vision DeepResearch~\cite{huang2026vision}, and Tongyi DeepResearch~\cite{team2025tongyi}. Implementation details, including training hyperparameters, inference hyperparameters, and compute resources, are provided in Appendix~\ref{appendix:training_inference_hyperparameters}.

\subsection{Main results}

\input{Tables/main_table}

    As shown in Table~\ref{tab:main_table}, we compare ToolAnchor-DeepResearch-30B-A3B against both proprietary and open-source deep research agents across three benchmarks.

    \paragraph{Visual search.} 
    
        On VDR-Bench, ToolAnchor achieves 28.8\%, outperforming Gemini-2.5-Pro (18.8\%), GPT-5 (20.4\%), and Qwen3-VL-30B-A3B-Thinking (23.2\%), despite using a smaller 30B LLM without prior visual-task training. This substantially narrows the gap with a vision-language model (VLM) agent trained from scratch for visual search, namely Vision-DeepResearch-30B-A3B (37.8\%). This performance gap is a predictable trade-off: Vision-DeepResearch is a native vision-language model (VLM) agent jointly optimized from scratch on a 5$\times$ larger multimodal dataset (15K instances). In contrast, ToolAnchor operates on a frozen, text-only LLM backbone augmented purely via dynamic tool APIs. Achieving a competitive 28.8\% without expensive multimodal alignment or massive data dependency underscores the remarkable parameter and compute efficiency of our toolset expansion paradigm.
    
    \paragraph{General assistant and textual search.} 
    
        On GAIA, ToolAnchor achieves 74.5\%, improving over the base model (70.9\%) by 3.6 percentage points and surpassing GPT-5 (67.4\%), Claude-4-Sonnet (68.3\%), and OpenAI DeepResearch (67.4\%). On BrowseComp, ToolAnchor achieves 45.0\%, a 1.6 percentage point improvement over the base model (43.4\%). These results demonstrate that ToolAnchor enables the agent to acquire new visual tool-use capabilities while largely preserving its performance on the original textual tasks. The slight improvement observed on GAIA and BrowseComp suggests that internalizing effective counterfactual anchor contexts can also reinforce the agent’s existing tool usage and reasoning policies.

\subsection{Ablation study}

    \paragraph{Visual tools.}

        Table~\ref{tab:visual_tools} compares the baseline Tongyi-DeepResearch-30B-A3B model with a variant equipped with our visual tools. The results demonstrate that while the integration of these visual tools significantly boosts performance on the VDR-Bench visual search (from 7.5\% to 16.6\%), it incurs a marginal performance degradation on textual search benchmarks (GAIA and BrowseComp).

\input{Tables/visual_tools_ablation_study}

    \paragraph{Post-training method.}

\begin{figure}[t]
            \centering
            \begin{minipage}{0.49\textwidth}
                \centering
                \includegraphics[width=\textwidth]{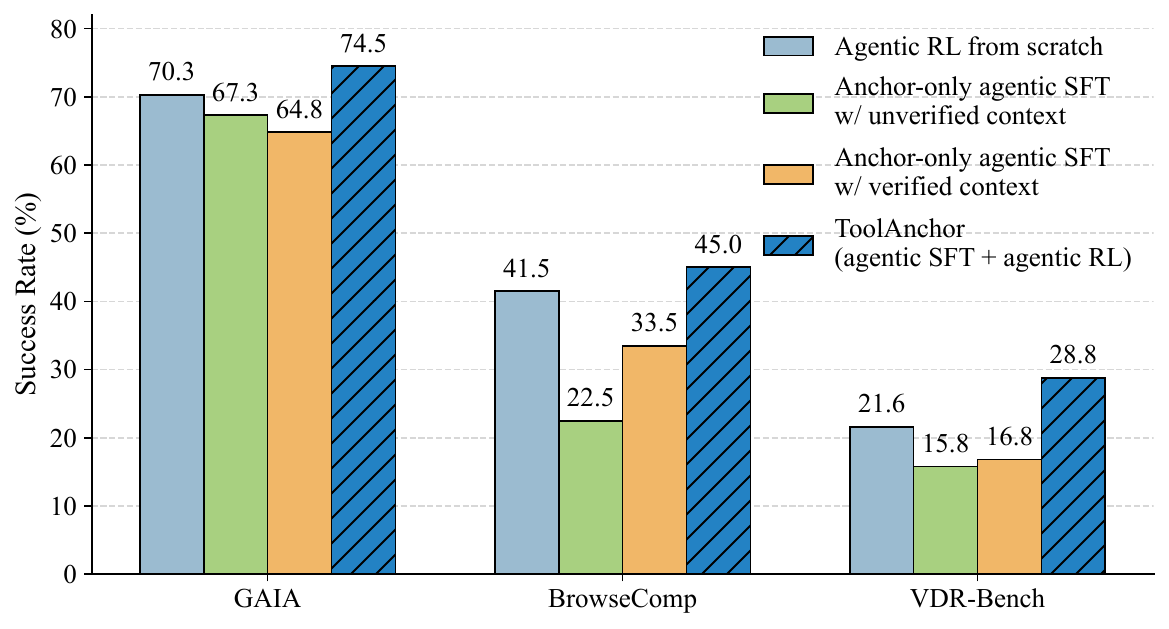}
                \captionof{figure}{Effects of post-training components}
                \label{fig:post_train_ablation_study}
            \end{minipage}
            \hfill
            \begin{minipage}{0.49\textwidth}
                \centering
                \includegraphics[width=\textwidth]{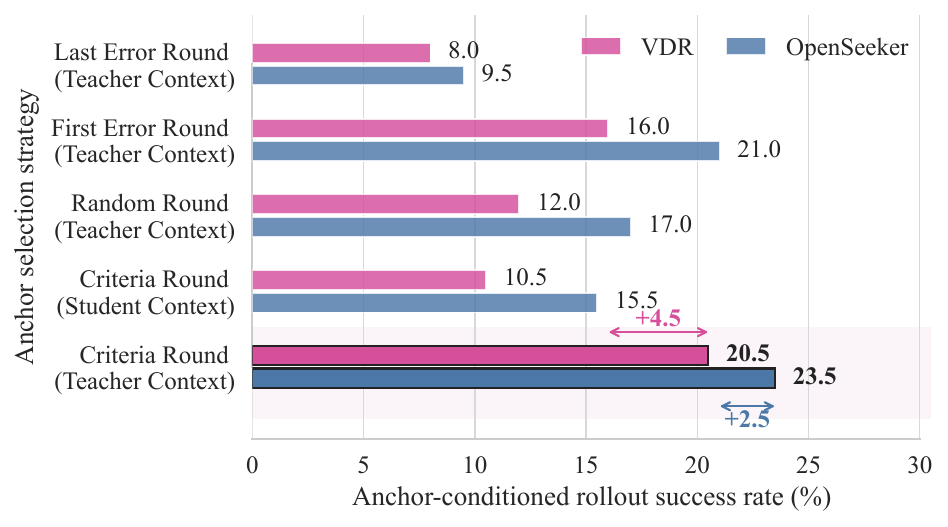}
                \vspace{-14pt}
                \captionof{figure}{\small Effects on anchor round selection strategies}
                \label{fig:anchor_selection_strategies}
            \end{minipage}
        \end{figure}

        Figure~\ref{fig:post_train_ablation_study} ablates the post-training pipeline after visual-tool expansion. Agentic SFT on unverified counterfactual contexts performs poorly, especially on BrowseComp, indicating that teacher-hypothesized contexts can introduce noisy or ineffective supervision if applied without student-side verification. Using verified contexts improves over unverified agentic SFT on BrowseComp and VDR-Bench, supporting the utility of counterfactual anchor context verification. Agentic RL trained from scratch achieves moderate performance (70.3\%, 41.5\%, 21.6\%), indicating that without access to counterfactual anchor contexts via SFT, RL alone exhibits limited exploration and fails to fully leverage the available toolset. However, agentic SFT with verified contexts alone still underperforms the ToolAnchor, suggesting that anchor-context internalization through SFT is insufficient to fully recover and improve the agent's tool-use policy. By combining verified SFT with a subsequent agentic RL, ToolAnchor achieves the best performance across all three benchmarks. This indicates that verified counterfactual contexts provide effective local supervision, while the post-SFT RL stage aligns the internalized anchor behaviors with the agent's previously learned policy.

\subsection{Further analysis and discussions}

\begin{figure}
            \centering
            \includegraphics[width=1.0\linewidth]{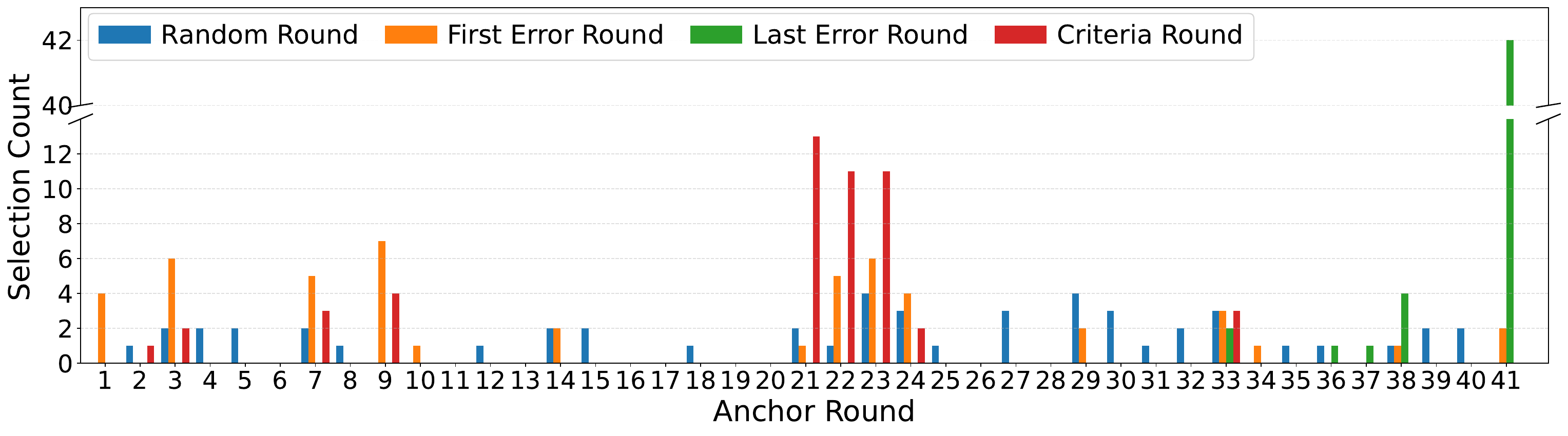}
            \caption{Distribution of anchor rounds over 50 repeated teacher hypotheses for a failed trajectory.}
            \label{fig:anchor_round_distribution}
        \end{figure}

    \paragraph{Does the choice of anchor round matter?} To isolate the efficacy of our criterion-based anchor selection from the mere presence of teacher intervention, we benchmark it against three heuristic baselines: intervening at a random round, the first erroneous round, and the last erroneous round. Evaluated across 200 instances from both OpenSeeker and VDR-Train (using Gemini-3.1-Pro-Preview as the teacher), our criterion-based strategy consistently yields the highest recovery rates. We repeat teacher anchor selection 50 times on one failed trajectory. As illustrated in Figure~\ref{fig:anchor_round_distribution}, effective anchors heavily cluster around strategic causal pivots rather than trivial first/last-error boundaries, confirming that a precise anchor location is as critical as the counterfactual context itself.

    \paragraph{How many counterfactual contexts survive verification?}

        For each teacher-generated counterfactual context, we sample three student rollouts and use Pass@3 task-level verification, where a context passes if any rollout solves the task. We then apply anchor-level verification to successful rollouts, retaining only contexts whose success is judged materially attributable to the anchor intervention. Overall, 33.67\% of candidates pass task-level verification and 31.98\% remain after anchor-level verification, showing that verification filters noisy teacher hypotheses while retaining contexts that both recover the task and validly intervene on the failure point.

\subsection{Case study}

    In Figure~\ref{fig:toolanchor_teaser_figure}, we present a case study to illustrate how ToolAnchor redirects a failed visual-search trajectory. In this example, the base Tongyi DeepResearch Agent first uses visual tools to locate the green-hooded character and invoke the crop-and-image search tool, but then dismisses the returned visual-search leads as generic and switches to broad text search, eventually producing an incorrect answer. The teacher hypothesizes that this decision is the counterfactual anchor round, since it is the key point where the agent abandons the visual-evidence path. The generated counterfactual context instead keeps the agent on the visual-search path by analyzing the crop-and-image search results and issuing a targeted follow-up query. This qualitative example illustrates how a local counterfactual context can redirect the subsequent rollout. Details are provided in Appendix~\ref{appendix:case_study}. Consistent with this qualitative example, Appendix~\ref{appendix:counterfactual_anchor_rounds} shows that hypothesized counterfactual anchor rounds are broadly distributed across different trajectory positions, with most occurring within the first 10-20 rounds, and that each can potentially produce distinct effects on the agent's subsequent behavior.

%% file: Tables/main_table.tex
\begin{table*}[htbp!]
    \centering
    \caption{Comparison of ToolAnchor and baseline deep research agents across benchmarks. Methods marked with * indicate that their results are taken from the original or reproduced papers (see Appendix~\ref{appendix:tool_description} for toolset details).}
    \small
    \resizebox{\textwidth}{!}{
    \begin{tabular}{lllccc}
        \toprule
        \textbf{Category} & \textbf{Method} & \textbf{Tools} & \textbf{GAIA} & \textbf{BrowseComp} & \textbf{VDR-Bench} \\
        \midrule
        \multirow{3}{*}{\textbf{LLM Agent}} 
        & OpenAI DeepResearch & Tongyi & 67.4\%* & 51.5\%* & - \\
        & OpenSeeker-v1-30B-SFT & OpenSeeker & - & 29.5\% & -  \\
        & Tongyi-DeepResearch-30B-A3B & Tongyi & 70.9\%* & 43.4\%* & 7.5\%* \\ 
        \midrule
        \multirow{6}{*}{\textbf{MLLM Agent}}
        & GPT-5 & Tongyi+VDR & 67.4\%* & 51.5\%* & 20.4\%* \\
        & Gemini-2.5-Pro & Tongyi+VDR & 74.5\% & 10.0\% & 18.8\%* \\
        & Claude-4-Sonnet & Tongyi+VDR & 68.3\%* & 12.2\%* & 13.6\%* \\
        & Qwen3-VL-30B-A3B-Thinking & Tongyi+VDR & 42.4\% & 5.0\% & 23.2\%* \\
        & Qwen3-VL-30B-A3B-Instruct & Tongyi+VDR & 32.7\% & 4.5\% & 20.2\%* \\
        & Vision-DeepResearch-30B-A3B & Tongyi+VDR & 55.2\% & 18.5\% & 37.8\%* \\
        \midrule
        \textbf{Ours}
        & ToolAnchor-DeepResearch-30B-A3B & Tongyi+Ours & 74.5\% & 45.0\% & 28.8\% \\
        \bottomrule
    \end{tabular}
    \label{tab:main_table}
    }
\end{table*}

%% file: Tables/visual_tools_ablation_study.tex
\begin{table*}[htbp!]
    \centering
    \caption{Effect of visual tool integration on performance across benchmarks}
    \small
    \begin{tabular}{llcc}
        \toprule
        Method & GAIA & BrowseComp & VDR-Bench  \\
        \midrule
        Tongyi-DeepResearch-30B-A3B & 70.9\%* & 43.4\%*  & 7.5\%* \\ 
        Tongyi-DeepResearch-30B-A3B w/ visual tools & 70.3\% & 42.5\%  & 16.6\% \\ 
        \bottomrule
    \end{tabular}
    \label{tab:visual_tools}
\end{table*}

%% file: Sections/conclusion.tex
In this paper, we propose ToolAnchor, a framework for extending post-trained tool-use agents with new tools through verified counterfactual context anchoring. Our main insight is that toolset expansion is not merely about adapting tool-selection and reasoning patterns to newly introduced tools. Post-trained agents may also need effective contextual anchors to break out of established reasoning and tool-use patterns. ToolAnchor turns teacher-hypothesized counterfactual contexts into training signals only after student-side verification, allowing the agent to internalize local high-leverage redirections rather than imitate full teacher trajectories. Experiments across general assistance, textual search, and visual search show that this mechanism improves new visual-tool use while also benefiting existing tool use and agentic reasoning, suggesting a practical path for capability expansion without retraining agents from scratch. In this work, the experiments rely on a limited set of teacher models and verified anchor contexts. Future research could investigate incorporating a more diverse set of teacher models and expanding the training dataset to enhance coverage, robustness, and generalization of internalized behaviors across a broader range of tasks and toolsets.

%% file: appendix.tex
\section{Detailed case study}
\label{appendix:case_study}

    We first present a failed inference trajectory from the Tongyi DeepResearch Agent in Appendix~\ref{appendix:deepresearch_agent_inference_case}, with the corresponding agent prompt provided in Appendix~\ref{appendix:prompt_deep_research_agent}. We then show the teacher model's counterfactual anchor-round hypothesis for this trajectory in Appendix~\ref{appendix:anchor_hypothesis_case}, with the corresponding hypothesis prompt provided in Appendix~\ref{appendix:prompt_anchor_hypothesis}. Finally, we provide the counterfactual anchor context generated for the hypothesized anchor round in Appendix~\ref{appnedix:continual_generation_case}, with the corresponding counterfactual context generation prompt provided in Appendix~\ref{appendix:prompt_anchor_hypothesis}.

    \subsection{Failed inference trajectory of the deep research agent}
    \label{appendix:deepresearch_agent_inference_case}

        \input{Appendix/case_study_trajectory}

    \subsection{Teacher counterfactual anchor round hypothesis}
    \label{appendix:anchor_hypothesis_case}

\input{Appendix/case_study_anchor}

    \subsection{Teacher counterfactual anchor context generation}
    \label{appnedix:continual_generation_case}

\input{Appendix/case_study_continual_generation}

\section{Prompt templates}
\raggedbottom

\subsection{System prompt for deep research agent}
\label{appendix:prompt_deep_research_agent}

    Following the Tongyi DeepResearch Agent, we provide the system prompt used by ToolAnchor, with the original tool-use instructions extended to include the new visual tools.

    \input{Appendix/prompt_deep_research_agent}

\subsection{System Prompt for teacher model counterfactual anchor round hypothesis}
\label{appendix:prompt_anchor_hypothesis}

    We provide the teacher prompts for counterfactual anchor round hypothesis and counterfactual anchor context generation.

\input{Appendix/prompt_counterfactual_hypothesis}

\input{Appendix/prompt_first_error_hypothesis}

\input{Appendix/prompt_last_error_hypothesis}

\input{Appendix/prompt_continual_generation}

\subsection{Prompt for anchor-level LLM-As-Judge verification}

    We provide the anchor-level LLM-as-judge verification prompt below.

    \input{Appendix/prompt_anchor_level_judge}

\subsection{Prompt for LLM-As-Judge evaluation}
\label{appendix:prompt_evaluation}

    For GAIA, BrowseComp, and VDR-Bench, we use the official evaluation prompts provided by each benchmark.

    \input{Appendix/prompt_benchmark_evaluation}

\section{ToolAnchor details}

    \subsection{Tool descriptions}
    \label{appendix:tool_description}

\input{Appendix/tool_description}

    \subsection{Training and inference details}
    \label{appendix:training_inference_hyperparameters}

\input{Appendix/training_inference_details}

    \subsection{Statistical analysis of counterfactual anchor rounds}
    \label{appendix:counterfactual_anchor_rounds}

\begin{figure}[htbp]
            \centering
            \begin{subfigure}[b]{0.48\textwidth}
                \centering
                \includegraphics[width=\textwidth]{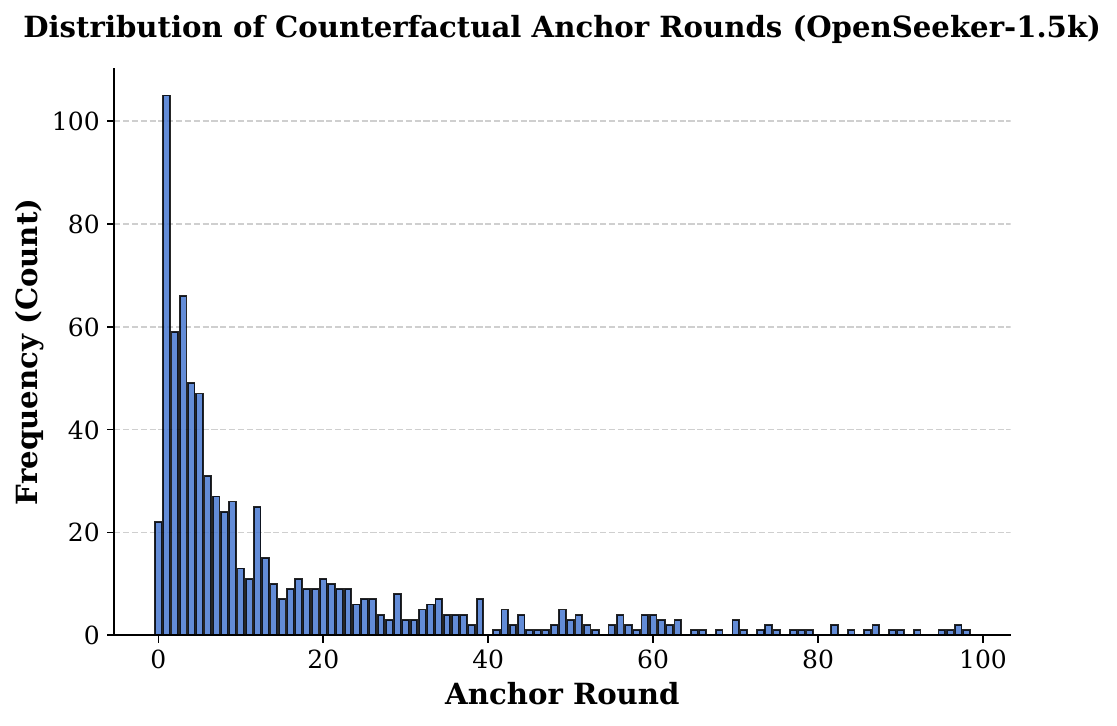}
            \end{subfigure}
            \hfill 
            \begin{subfigure}[b]{0.48\textwidth}
                \centering
                \includegraphics[width=\textwidth]{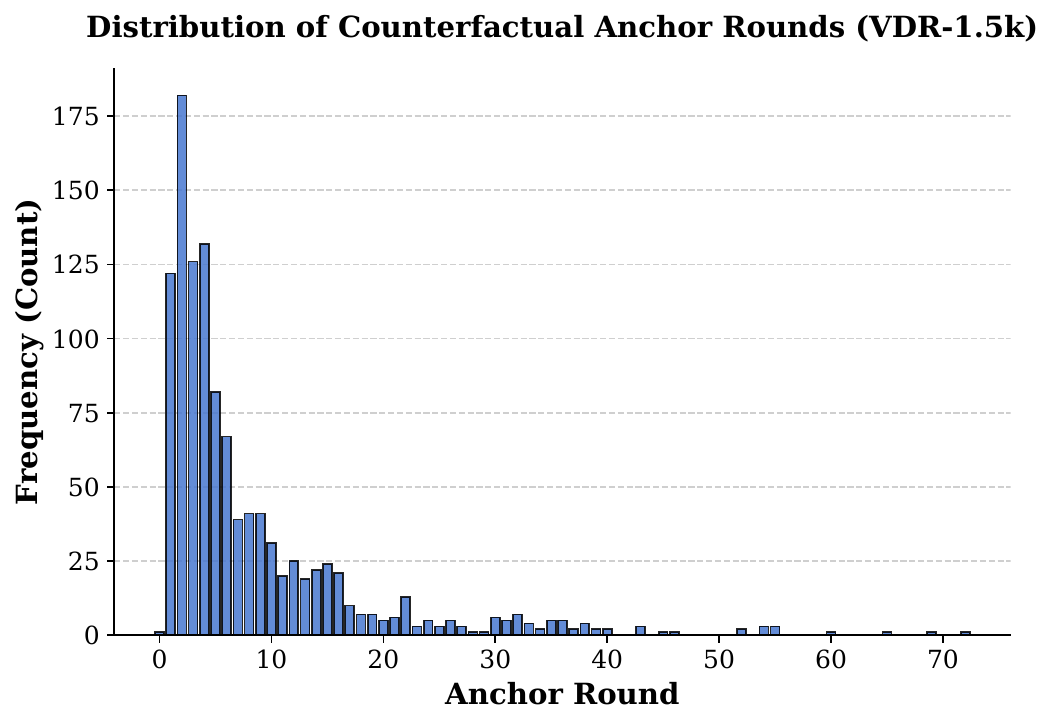}
            \end{subfigure}
            \caption{Frequency Statistics of Counterfactual Anchor Rounds in Gemini-2.5-Pro}
            \label{fig:statistics_analysis_conterfactual_anchor_rounds}
        \end{figure}

        Figure~\ref{fig:statistics_analysis_conterfactual_anchor_rounds} shows the distribution of counterfactual anchor rounds hypothesized by Gemini-2.5-Pro on OpenSeeker-1.5k and VDR-1.5k. On both datasets, the hypothesized anchor rounds are heavily concentrated in the early stages of trajectories, with most anchors occurring within the first 10--20 rounds. This suggests that early counterfactual contexts exert greater influence on subsequent reasoning and tool-use decisions, while preserving sufficient rollout horizon for exploration and recovery. Similarly, on VDR-1.5k, the early-round concentration is even more pronounced, indicating that visual-search tasks particularly benefit from introducing visual-evidence-oriented contexts near the beginning of the trajectory. The long-tail of the distribution indicates that valuable counterfactual contexts also occur at later rounds in some trajectories. These observations support our use of teacher-guided anchor-round hypothesis generation rather than relying on fixed heuristics such as always selecting the first round.

\section{Data contamination}

    To mitigate search-time data contamination~\cite{han2025search}, we filter out specific domain names, such as github.com and huggingface.co, during the web search process.

%% file: Appendix/case_study_trajectory.tex
\begin{tcolorbox}[
  title=Tongyi Deep Research Agent Case Study,
  breakable
]

(Uploaded 1 file: ['hard1\_Game\_115\_1.jpg'])

\begin{center}
    \includegraphics[width=1.0\linewidth]{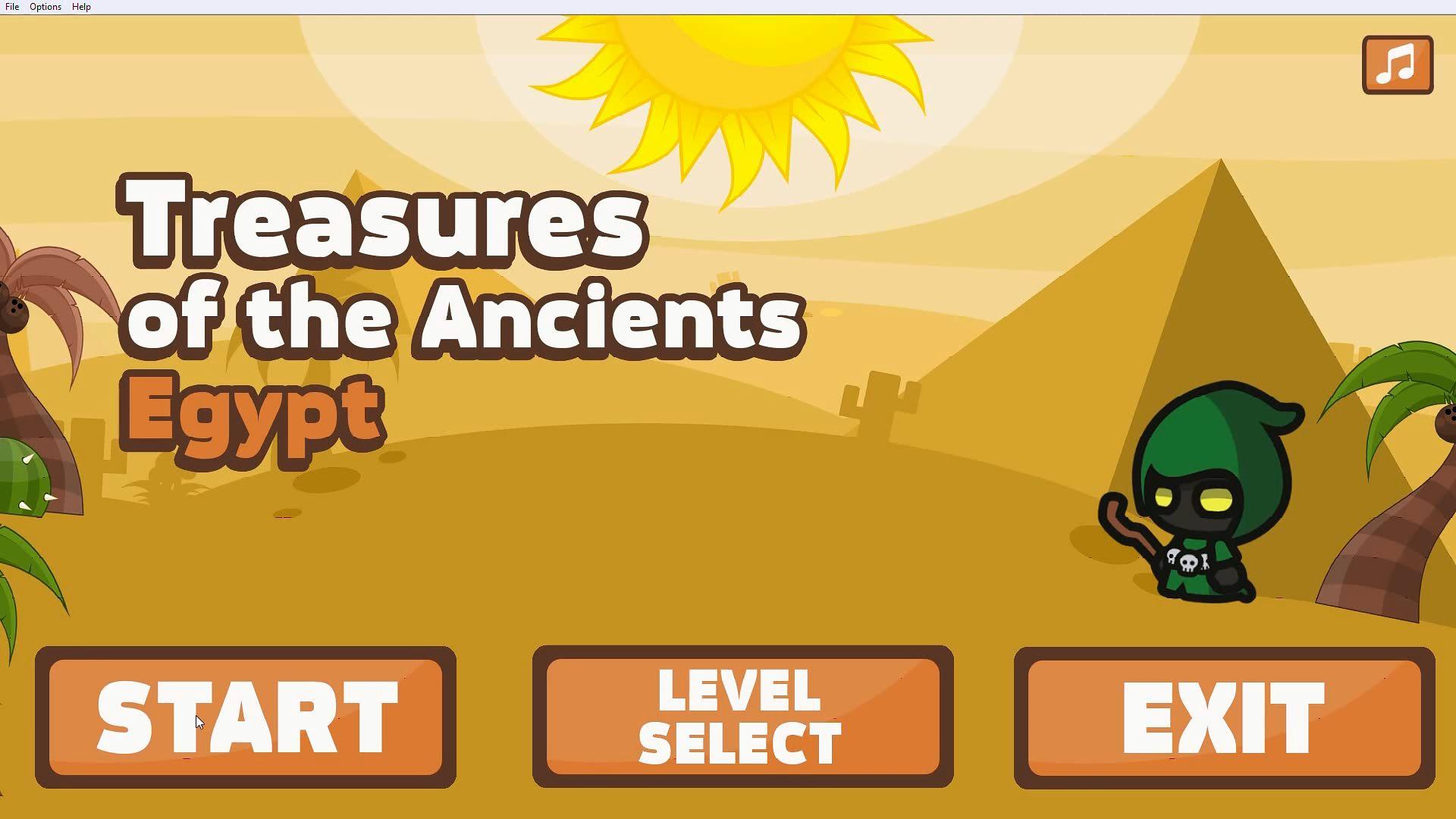}
\end{center}

The character in the green hood is from a game developed by a studio that also released a VR title in 2017. What is the name of the development studio, and what is the title of their 2017 VR game?

\vspace{0.1em}
\begin{tcolorbox}[colback=gray!60, colframe=gray!60, sharp corners, boxrule=0pt, boxsep=2pt, left=0pt, right=0pt, top=2pt, bottom=2pt, halign=center, width=\linewidth]
\textbf{Step 1}
\end{tcolorbox}
\vspace{0.1em}

<think>\\
We have an uploaded image: hard1\_Game\_115\_1.jpg. The user asks: "The character in the green hood is from a game developed by a studio that also released a VR title in 2017. What is the name of the development studio, and what is the title of their 2017 VR game?" So we need to identify the character in the green hood, find which game they are from, then find which studio developed that game, then find that studio's VR title released in 2017.\\
\\
We need to analyze the image to see the character. Let's use parse\_file to get some description or maybe OCR? We can use the visit tool after extracting from the image? Since it's an uploaded file, we need to parse file. Use parse\_file on 'hard1\_Game\_115\_1.jpg'. But parse\_file likely extracts text (OCR) but not visual analysis. However, we may need to view the image content. The tool does not show the image directly; maybe the parse\_file returns something like alt text? Actually parse\_file tool is for analyzing uploaded local files such as PDF, DOCX, etc. It may extract text from images via OCR.\\
\\
Given that the image is likely a screenshot from a game showing a character with a green hood. Could be multiple possibilities.\\
\\
Let's try parse\_file.\\
</think>\\
\\
<tool\_call>\\
\{"name": "parse\_file", "arguments": \{"files": ["hard1\_Game\_115\_1.jpg"]\}\}\\
</tool\_call>\\
\\
<tool\_response>\\
\relax['File token number: 7 File content:', '\# Error processing hard1\_Game\_115\_1.jpg: Document parsing failed']\\
</tool\_response>

\vspace{0.1em}
\begin{tcolorbox}[colback=gray!60, colframe=gray!60, sharp corners, boxrule=0pt, boxsep=2pt, left=0pt, right=0pt, top=2pt, bottom=2pt, halign=center, width=\linewidth]
\textbf{Step 2}
\end{tcolorbox}
\vspace{0.1em}

\textless{}think\textgreater{}\\{}
The parse\_file tool fails to process the image: "Document parsing failed". So we need another approach. Possibly we can use a reverse image search via bbox extractor and crop\_and\_search. The image likely contains a recognizable character with a green hood. Could be from a game like "The Legend of Zelda: Breath of the Wild"? There's Link wearing a green tunic, but not a hood specifically. A green hood could be from "The Witcher 3"? Not sure.\\{}
\\{}

Let's try using bbox\_extractor to detect relevant entities in the image. The function requires an image\_name and a query. The query could be something like "character wearing green hood". Let's try.\\{}
\textless{}/think\textgreater{}\\{}
\\{}
\textless{}tool\_call\textgreater{}\\{}
\{"name": "bbox\_extractor", "arguments": \{"image\_name": "hard1\_Game\_115\_1.jpg", "query": "character wearing green hood"\}\}\\{}
\textless{}/tool\_call\textgreater{}

\textless{}tool\_response\textgreater{}\\{}
\{"bbox": [[757, 469, 902, 744]]\}\\{}
\textless{}/tool\_response\textgreater{}

\vspace{0.1em}
\begin{tcolorbox}[colback=gray!60, colframe=gray!60, sharp corners, boxrule=0pt, boxsep=2pt, left=0pt, right=0pt, top=2pt, bottom=2pt, halign=center, width=\linewidth]
\textbf{Step 3}
\end{tcolorbox}
\vspace{0.1em}

\textless{}think\textgreater{}\\{}
We got a bounding box: [[757, 469, 902, 744]]. That suggests there is a region in the image containing the character wearing green hood. Let's crop and search that region.\\{}
\\{}
We'll call crop\_and\_search with that bbox.\\{}
\textless{}/think\textgreater{}\\{}
\\{}
\textless{}tool\_call\textgreater{}\\{}
\{"name": "crop\_and\_search", "arguments": \{"image\_name": "hard1\_Game\_115\_1.jpg", "bbox": [757, 469, 902, 744]\}\}\\{}
\textless{}/tool\_call\textgreater{}

\textless{}tool\_response\textgreater{}\\{}
Google Lens results for bbox [757, 469, 902, 744] in image url https://agent-tool-use.oss-cn-beijing.aliyuncs.com/crop\_hard1\_Game\_115\_1\_\_\_757\_469\_902\_744.jpg (top 10):\\{}
\\{}
1. [Wizard Tiny Style 2D Character Sprites - CraftPix.net](Web page URL: https://craftpix.net/product/wizard-tiny-style-2d-character-sprites/)\\{}
Source: CraftPix\\{}
The original query image and the search result image both feature a cartoonish character with green hair, yellow eyes, and a black mask holding a staff. However, there are subtle differences such as the background color and the presence of text in the search result image. Despite these minor variations, the core design elements and overall appearance of the character are consistent between the two. Therefore, they depict the Exact Same Entity with the primary visual reason being the identical character design and features that match precisely across both images.\\{}
Image URL: https://encrypted-tbn3.gstatic.com/images?q=tbn:ANd9GcQ6njpKxAOOTmg8j\\8x3T3zMvyqgA59L7ZuZMaH6qk7S8wN-P3W4\\{}
\\{}
2. [Discover 12 enemies for games and skeleton characters in games ideas | skeleton warriors cartoon, 2d game rock sprites, skeleton game character and more](Web page URL: https://www.pinterest.com/giorgoschar06/enemies-for-games/)\\{}
Source: Pinterest\\{}
The original query image and the search result image share several visual similarities including a character with green hair, a black mask, and a skull emblem on its outfit both set against a simple background However the original query image features a single character with more detailed and vibrant colors while the search result image includes three characters with less detailed and more muted colors Despite these differences the central character in the search result image closely matches the one in the original query image leading to the conclusion that they depict an Exact Same Entity primarily because the distinctive features like the green hair black mask and skull emblem are consistent between the two images.\\{}
Image URL: https://encrypted-tbn0.gstatic.com/images?q=tbn:ANd9GcTtGe3MHCcQf5-Z0NLydwHSLvYpwVGUOSCq8YmKUMZU\_upmR7fF\\{}
\\{}
3. [Before The Halloween Update All Animals Evolution With Patrick Reaper Ghostly Pumpkin (EvoWorld.io) - YouTube](Web page URL: https://www.youtube.com/watch?v=ewpDuL2JsB0)\\{}
Source: YouTube\\{}
The original query image and the webpage image share several visual similarities including a character with green hair, yellow eyes, and a black mask but differ in context and additional elements. The character in the original query image appears alone while the webpage image shows multiple characters, including the one that resembles the original, in a more complex scene with text and other animals. Despite these differences, the central character's design is nearly identical suggesting they are the Exact Same Entity. The primary visual reason for this conclusion is the consistent depiction of the character's distinctive features such as the green hair and yellow eyes which match precisely between both images.\\{}
Image URL: https://encrypted-tbn3.gstatic.com/images?q=tbn:ANd9GcQ7Y1DEdHlp9BRiwl\\
z8tpA9Gb39EGJYdKptXOrIvcVFGmsow\_ur\\{}
\\{}
4. [Skins | EvoWorld.io Wiki | Fandom](Web page URL: https://evoworldio.fandom.com/wiki/Skins)\\{}
Source: Fandom\\{}
The original query image and the search result image both feature cartoon characters wearing green hooded outfits but exhibit distinct differences in design and context. The character in the original query image has black facial features with yellow eyes and holds a scythe while the character in the search result image has a brown face with green eyes and holds a different type of weapon. These differences in facial features, eye color, and weaponry indicate that they do not represent the Exact Same Entity but rather Similar/Related Entities. The primary visual reason for this conclusion is the variation in the characters' physical attributes and accessories which suggest they are separate though related figures.\\{}
Image URL: https://encrypted-tbn1.gstatic.com/images?q=tbn:ANd9GcSyY8HC7cNk1mR-6KgR88z7eQ3oPAsVdlhKODuXuxnddLqD2B2X\\{}
\\{}
5. [Kaden | The Rewrittenverse Wiki | Fandom](Web page URL: https://the-rewrittenverse.fandom.com/wiki/Kaden)\\{}
Source: Fandom\\{}
The original query image and the search result image both feature a character wearing a green hooded mask with glowing yellow eyes and a skull emblem on their chest indicating a thematic similarity. However, there are notable differences such as the background color, the presence of a sword in the search result image, and slight variations in the character's posture and design elements like the mask's shape. Despite these differences, they appear to represent the Exact Same Entity as the core visual identity and design elements are consistent. The primary visual reason for this conclusion is the matching distinctive green hooded mask and skull emblem which are central to the character's appearance in both images.\\{}
Image URL: https://encrypted-tbn2.gstatic.com/images?q=tbn:ANd9GcTehYBVDpN-k5QthqtlieesGUVTmNFhtjIKFxuP4cpxzL9lXSg0\\{}
\\{}
6. [North-East Europe Is Kinda Toxic | Evoworld - YouTube](Web page URL: https://www.youtube.com/watch?v=\_kGnzn0mUVg)\\{}
Source: YouTube\\{}
The original query image and the search result image both feature a character with green hair, yellow eyes, and a black mask which suggests a strong visual connection. However, the original query image shows a single character against a plain background while the search result image includes multiple characters in a more complex scene with additional elements like water, skulls, and text indicating "North-East Europe." Despite these differences, the central character appears identical in design and style. Therefore, they depict the Exact Same Entity with the primary visual reason being the consistent and unique design of the main character across both images.\\{}
Image URL: https://encrypted-tbn2.gstatic.com/images?q=tbn:ANd9GcRXAhITalNodtLDc\\
EKvMpjeQOBAxNAnJNNbLgrtFIhZ3tEGg7k3\\{}
\\{}
7. [Grinding to Rank \#1 in West Europe 1 | EvoWorld Gameplay Live  - YouTube](Web page URL: https://www.youtube.com/watch?v=JRxO\_19M1gY)\\{}
Source: YouTube\\{}
The original query image and the search result image both feature a character with green hair, yellow eyes, and a black mask but the original query image shows a single character while the search result image includes multiple characters in a different setting with additional elements like text and other figures. Despite these differences, the central character appears visually consistent in design and style. They depict a Similar/Related Entity because the primary visual reason for this conclusion is that the main character shares identical distinctive features such as the green hair and yellow eyes which are unique identifiers despite being presented in different contexts.\\{}
Image URL: https://encrypted-tbn1.gstatic.com/images?q=tbn:ANd9GcTWZ-Ho\_PyRunSooVQQn8Q6R5jhYkT4a9SBOiuFRn0BJpfYYdwX\\{}
\\{}
8. [Green Witcher | Raid Heroes Wiki | Fandom](Web page URL: https://raid-heroes.fandom.com/wiki/Green\_Witcher)\\{}
Source: Fandom\\{}
The original query image and the search result image both feature a cartoon character wearing a green hooded outfit with a mask but differ in several details such as eye color, additional accessories like a skull emblem, and background colors. Despite these differences, the core design elements including the pose, clothing style, and overall silhouette are remarkably similar suggesting a strong connection between the two. They depict a Similar/Related Entity primarily because while there are distinct variations in the details, the fundamental character design remains consistent across both images.\\{}
Image URL: https://encrypted-tbn2.gstatic.com/images?q=tbn:ANd9GcSQamudfIe3m0pWox\\
WH1sI4PgmkSg3o7gIWoMDwBvhLrMVmSsvY\\{}
\\{}
9. [GEE estimates of negative binomial models for number of first time... | Download Table](Web page URL: https://www.researchgate.net/figure/GEE-estimates-of-negative-binomial-models-for-number-of-first-time-entries-in-the-segment\_tbl1\_237290427)\\{}
Source: ResearchGate\\{}
The original query image and the search result image share striking similarities including a character with green hair, yellow eyes, a black mask, and a skull symbol on its chest both set against contrasting backgrounds. However, there are subtle differences such as the background color and the angle at which the character is depicted. Despite these minor variations, the core design elements are consistent leading to the conclusion that they depict the Exact Same Entity. The primary visual reason for this conclusion is the identical character design and features which are unmistakably the same despite slight contextual differences.\\{}
Image URL: https://encrypted-tbn1.gstatic.com/images?q=tbn:ANd9GcRpc\_u3xbZUbFUEK\\
1SA9KxFHBPvvY8HAjRRaIDYAfD4qIyxxPOe\\{}
\\{}
10. [Cloaked Plant Wizard | Prodigy Game Wiki | Fandom](Web page URL: https://prodigy-game.fandom.com/wiki/Cloaked\_Plant\_Wizard)\\{}
Source: Fandom\\{}
The original query image and the search result image share similar thematic elements such as a character wearing a green hooded outfit but differ significantly in design and context. The original query image features a cartoonish figure with glowing yellow eyes and a skull emblem, while the search result image depicts a more detailed and realistic character holding a weapon. Despite these differences, both characters wear green hoods and have a dark face mask suggesting a connection. They depict a Similar/Related Entity rather than the Exact Same Entity primarily because the overall design, accessories, and artistic style are distinct enough to indicate different representations of a similar concept.\\{}
Image URL: https://encrypted-tbn2.gstatic.com/images?q=tbn:ANd9GcTcGmi-TZlxRrtrW35zinLEEEQ4-RANZDDXthAr8ZRiLKDAJ7EJ\\{}
\textless{}/tool\_response\textgreater{}

{\color{red}

\vspace{0.1em}
\begin{tcolorbox}[colback=gray!60, colframe=gray!60, sharp corners, boxrule=0pt, boxsep=2pt, left=0pt, right=0pt, top=2pt, bottom=2pt, halign=center, width=\linewidth]
\color{red}
\textbf{Step 4}
\color{black}
\end{tcolorbox}
\vspace{0.1em}

\textless{}think\textgreater{}\\{}
The search returned many results, but they don't directly identify a known game character. They refer to "Wizard Tiny Style 2D Character Sprites", "EvoWorld.io" etc. Might be generic sprites.\\{}
\\{}
Thus the image may depict a character from an indie game perhaps with green hood.\\{}
\\{}
Let's try different query: maybe the character is "Fenix"? No.\\{}
\\{}
Better approach: We need to see the image content. Since parse\_file fails and crop\_and\_search yields generic results, we need to infer manually based on the context of typical game characters with green hood.\\{}
\\{}
\\{}
}

{\color{red}
The question: "The character in the green hood is from a game developed by a studio that also released a VR title in 2017. What is the name of the development studio, and what is the title of their 2017 VR game?" So we need the development studio of the game featuring this character. The studio also released a VR title in 2017. Which studios have VR titles in 2017? There are many: Ubisoft (Star Trek: Bridge Crew), Oculus Studios (Lone Echo?), though Lone Echo is 2017 by Ready At Dawn (published by Sony). Actually "Lone Echo" was 2017 by Ready at Dawn Studios, published by Sony Interactive Entertainment; it's a VR game.\\{}
\\{}

Other VR releases 2017 include:\\{}
\\{}
- "Beat Saber" (by Beat Games)\\{}
- "Superhot VR" (Superhot Team)\\{}
- "Moss" (by Polyarc)\\{}
- "Arizona Sunshine" (by Vertigo Games)\\{}
- "Boneworks" (by Stress Level Zero)\\{}
- "Skyrim VR" (Bethesda Game Studios, but originally 2016, not 2017)\\{}
- "Until Dawn: Rush of Blood" (Supermassive Games)\\{}
}
{\color{red}
- "Resident Evil 7: Biohazard" (Capcom) but not VR exclusive.\\{}
\\{}
- "The Walking Dead: Saints \& Sinners" (Skydance Interactive, but release later).\\{}

- "Wolfenstein: The New Order VR"? Not sure.\\{}
\\{}
Also "Doom VFR" (id Software) maybe earlier.\\{}
\\{}
But which game developer made a notable green-hooded character? Many indie games have such characters.\\{}
\\{}
Alternatively, maybe the game is "Hades"? In Hades, there is a character wearing a green hood? Not exactly. Possibly "Hollow Knight"? Character like "Green Hood" but there is a "Hornet" in a green cloak maybe?\\{}
\\{}
Could be "Darkest Dungeon"? No.\\{}
\\{}
Maybe it's "Genshin Impact"? Character "Klee" has green hair/hod? But not a hood.\\{}
\\{}
There is a known game "Horizon Zero Dawn"? Character "Aloy" wears a small green hood? She has a hood but not green. Another is "Stardew Valley"? No.\\{}
\\{}
Think about games with distinctive green hooded characters: "Prince of Persia"? He wears a green tunic sometimes.\\{}
\\{}
Another thought: "Baten Kaitos"? Unlikely.\\{}
\\{}
Perhaps the screenshot is from "Dark Souls"? Some NPC wear green hoods, e.g., "Patches"? Not green.\\{}
\\{}
Could be "The Legend of Zelda: Ocarina of Time" where Link wears green tunic and cap, not hood though.\\{}
\\{}
Maybe it's "Octopath Traveler"? No.\\{}
\\{}
Let's examine the search results: Many references to "green hooded mask with glowing yellow eyes". That could be a specific enemy type in a game like "EvoWorld.io". But the screenshot might be from "EvoWorld.io"? However that is a web-based game. Does its developer also release a VR title in 2017? Not likely.\\{}
\\{}

Let's search for "green hood character video game" using search tool.\\{}
\textless{}/think\textgreater{}\\{}
\\{}
\textless{}tool\_call\textgreater{}\\{}
\{"name": "search", "arguments": \{"query": ["\textbackslash\{\}"green hood\textbackslash\{\}" video game character"]\}\}\\{}
\textless{}/tool\_call\textgreater{}
% \color{black}

\textless{}tool\_response\textgreater{}\\{}
A Google search for '"green hood" video game character' found 10 results:\\{}
% \\{}
}
{\color{red}
\#\# Web Results\\{}

1. [Green hood (Fallout 4) - Fallout Wiki](https://fallout.fandom.com/wiki/Green\_hood\_(Fallout\_\\
4))\\{}
The green hood is a piece of headwear that can be obtained in Fallout 4. The green hood provides a Damage Resistance of 1 and can be worn with a gas mask, ...\\{}

}
{\color{red}
\quad
}
{\color{red}

2. [Green Hood | IMO: The World of Magic Wiki - Fandom](https://imotwom.fandom.com/wiki/Green\_Hood)\\{}
\\{}
Green Hood A hooded fellow to help guide you. Pet Level Required: 8. Class Required: Magician-Class · Ranger-Class · Warrior-Class All Critical: 1\\{}
\\{}
3. [The Ultimater Guide to Green Hood! [3k Words!] | Hypixel Forums](https://hypixel.net/threads/the-ultimater-guide-to-green-hood-3k-words.1209267/)\\{}
Date published: Jul 5, 2017\\{}
\\{}
Cake Monster is a bad teammate for Hood, he likes to combo the hell out of everyone with insane DPS but has no way to work with Green Hood.\\{}
\\{}
4. [[TOMT] [GAME] very popular game where you play as a bald kid with ...](https://www.reddit.com/r/tipofmytongue/comments/16ff1qg/tomt\_gam\\
e\_very\_popular\_game\_where\_you\_play\_as\_a/)\\{}
Date published: Sep 10, 2023\\{}
\\{}
[TOMT] (anime character) blue hair green hood. 2. 4. Did ya see that ... [2007/2008][FLASH] Top Down game where the main character has a green ...\\{}
\\{}
5. [Smash Heroes: Green Hood - YouTube](https://www.youtube.com/watch?v=4gPONz90lrs)\\{}
Date published: Dec 23, 2016\\{}
\\{}
Happy 100th video! In this, I will be showing you guys how to use Green Hood effectively and to the best of his ability ... The Worst Character ...\\{}
\\{}
6. [Green Hood Guy Roblox - TikTok](https://www.tiktok.com/discover/green-hood-guy-roblox)\\{}
Date published: Feb 2, 2026\\{}
\\{}
5.8M posts. Discover videos related to Green Hood Guy Roblox on TikTok. See more videos about Roblox Green Guy, Who Is Green Guy with Green ...\\{}
\\{}
7. [Exploring the Green Hooded Avatars in Roblox - TikTok](https://www.tiktok.com/@creepiecrawlies6/video/7540798118984502542)\\{}
Date published: Aug 20, 2025\\{}
\\{}
... games, Roblox hype culture, comment your favorite Roblox game, Roblox player community ... Watch the hilarious video of a dog in a green hood ...\\{}
\\{}
8. [Smash Heroes] Minecraft - "*NEW* GREEN HOOD GAMEPLAY!"](https://www.youtube.com/watch?v=nhPS8UzR5\_k)\\{}
Date published: Aug 17, 2016\\{}
\\{}
... video description's due to a false positive on YouTube's end. As ... GREEN HOOD GAMEPLAY!" 3.2K views · 9 years ago. \#Christian ...more ...\\{}
\\{}
9. [Green hood (Fallout 76) | Fallout Wiki - Fandom](https://fallout.fandom.com/wiki/Green\_hood\_(Fallout\_76))\\{}
\\{}
The green hood is a piece of headwear that can be obtained in Fallout 76. This piece of headwear appears as a hood normally connected to the assault gas ...\\{}
\\{}
10. [Green Hood Character - Pinterest](https://www.pinterest.com/ideas/green-hood-character/907747242196/)\\{}
\\{}
}
{\color{red}
Discover Pinterest's best ideas and inspiration for Green hood character. Get inspired and try out new things. Last updated 4d.\\{}
\textless{}/tool\_response\textgreater{}

}

\vspace{0.1em}
\begin{tcolorbox}[colback=gray!60, colframe=gray!60, sharp corners, boxrule=0pt, boxsep=2pt, left=0pt, right=0pt, top=2pt, bottom=2pt, halign=center, width=\linewidth]
\textbf{Step 5}
\end{tcolorbox}
\vspace{0.1em}

\textless{}think\textgreater{}\\{}
Search results are not particularly helpful. Let's try search for "green hooded character game screenshot" maybe more specific.\\{}
\textless{}/think\textgreater{}\\{}
\\{}
\textless{}tool\_call\textgreater{}\\{}
\{"name": "search", "arguments": \{"query": ["\textbackslash\{\}"green hood\textbackslash\{\}" \textbackslash\{\}"character\textbackslash\{\}" \textbackslash\{\}"game\textbackslash\{\}" \textbackslash\{\}"screenshot\textbackslash\{\}""]\}\}\\{}
\textless{}/tool\_call\textgreater{}

\textless{}tool\_response\textgreater{}\\{}
A Google search for '"green hood" "character" "game" "screenshot"' found 10 results:\\{}
\\{}
\#\# Web Results\\{}
1. [What Does A Roblox Avatar with A Green Hood Green Skin ... - TikTok](https://www.tiktok.com/discover/what-does-a-roblox-avatar-with-a-green-hood-green-skin-and-round-head-mean)\\{}
Date published: Mar 2, 2026\\{}
\\{}
... character styles, detecting green hooded avatars, gaming ... Squid Game Roblox Avatar 0 Robux Whit Green Tshirt and Tshirt Screenshot.\\{}
\\{}
2. [[TOMT] (anime character) blue hai\\
r green hood - Reddit](https://www.reddit.com/r/tipofmytongue/comments/5l5zte/tomt\_\\textgreater{}anime\_character\_blue\_hair\_green\_hood/)\\{}
Date published: Dec 30, 2016\\{}
\\{}
[TOMT] (anime character) blue hair green hood. Solved. I can't ... screenshot that didn't have the bunny ears, and thought the lace was ...\\{}
\\{}
3. [How to get Green Bean outfit in Roblox \& Green Avatar explained](https://www.dexerto.com/roblox/how-to-get-green-bean-outfit-in-roblox-green-avatar-explained-3240530/)\\{}
Date published: Aug 20, 2025\\{}
\\{}
Grow a garden green hood ... keep reading. Ro Ghoul Roblox character in the diner. RobloxRo Ghoul codes (April 2026) · A screenshot featuring ...\\{}
\\{}
4. [Who Is Green Guy with Green Hoodie in Roblox - TikTok](https://www.tiktok.com/discover/who-is-green-guy-with-green-hoodie-in-roblox)\\{}
Date published: 5 days ago\\{}
\\{}
... Green Hood in Roblox, What Is The Green Hood ... Green Man Roblox stories, Roblox gameplay tips, Roblox character exploration, Roblox game ...\\{}
\\{}
5. [Berry avenue creature - Discover \& Share GIFs - Tenor](https://tenor.com/view/berry-avenue-creature-gif-3355296347644183555)\\{}
Date published: Oct 30, 2024\\{}
\\{}
Berry Avenue Creature Meme. a close up of a cartoon character with a green hood ... a screenshot of a video game with the words the audacity ...\\{}
\\{}
6. [My loot of last week Plus a green hood, because of rain and cold. A ...](https://www.instagram.com/p/DNN\_\_sesyj4/)\\{}
\\{}
Plus a green hood ... May be a video game screenshot. 17. chimerical\_dragonfly ... I create everything to complete a character and tell a story, from D\&D decor to ...\\{}
\\{}
7. [Roblox Green Avtsr Iwth Green Hood - TikTok](https://www.tiktok.com/discover/roblox-green-avtsr-iwth-green-hood)\\{}
Date published: Jan 5, 2026\\{}
\\{}
suggest character nga \#fyp \#roblox \#character ... Cara main Game PS2 di komputer kalian, cocok buat nostalgia game jadul kesukaan kalian nih.\\{}
\\{}
8. [Gods will fall. Wishlist now! https://f.mtr.cool/hdtqxfalbl - Instagram](https://www.instagram.com/reel/DWCW0mxjG7J/)\\{}
Date published: Mar 18, 2026\\{}
\\{}
A stylized illustration featuring a character in a green hood with glowing eyes and horns, ... Photo by Pupsker on March 09, 2026 ...\\{}
\\{}
9. [Makoto Naegi Danganronpa Sticker - Tenor](https://tenor.com/view/makoto-naegi-danganronpa-gif-4948538113774618876)\\{}
Date published: Jan 19, 2024\\{}
\\{}
Makoto Naegi Danganronpa Sticker. a stuffed doll with a green hood and a red zipper ... a screenshot of a video game with a character named ...\\{}
\\{}
10. [HD wallpaper: Fantasy, Warrior, Forest, Green, Hood 4096x2304px ...](https://www.wallpaperflare.com/fantasy-warrior-forest-green-hood-wallpaper-might/download/4096x2304)\\{}
\\{}
Fantasy, Warrior, Forest, Green, Hood ... warrior holding sword wallpaper, video game screenshot ... black haired girl anime character illustration ...\\{}
\textless{}/tool\_response\textgreater{}

\begin{center}
    \dots\dots\dots
\end{center}

\vspace{0.1em}
\begin{tcolorbox}[colback=gray!60, colframe=gray!60, sharp corners, boxrule=0pt, boxsep=2pt, left=0pt, right=0pt, top=2pt, bottom=2pt, halign=center, width=\linewidth]
\textbf{Step 91}
\end{tcolorbox}
\vspace{0.1em}

\textless{}think\textgreater{}\\{}
Thus not.\\{}
\\{}
Maybe the studio is "David Cseh" and they also made a non-VR game with a green hooded character. Could be a game like "Gates of Steem". But perhaps that game has a green hooded character? Let's search for Gates of Steem screenshot.\\{}
\textless{}/think\textgreater{}\\{}
\\{}
\textless{}tool\_call\textgreater{}\\{}
\{"name": "search", "arguments": \{"query": ["\textbackslash\{\}"Gates of Steem\textbackslash\{\}" \textbackslash\{\}"green\textbackslash\{\}""]\}\}\\{}
\textless{}/tool\_call\textgreater{}

\textless{}tool\_response\textgreater{}\\{}
A Google search for '"Gates of Steem" "green"' found 0 results:\\{}
\\{}
\#\# Web Results\\{}
\\{}
\textless{}/tool\_response\textgreater{}

\vspace{0.1em}
\begin{tcolorbox}[colback=gray!60, colframe=gray!60, sharp corners, boxrule=0pt, boxsep=2pt, left=0pt, right=0pt, top=2pt, bottom=2pt, halign=center, width=\linewidth]
\textbf{Step 92}
\end{tcolorbox}
\vspace{0.1em}

\textless{}think\textgreater{}\\{}
Thus not.\\{}
\\{}
Perhaps the green-hooded character is from a game called "The Last Stand". That is the answer we have. The studio is Con Artist Games, which also released a VR title in 2017: "VR Squash 2017"? But that seems mismatched. However maybe Con Artist Games contributed to VR Squash? Could be that David Cseh is associated with Con Artist Games? Let's search for "Con Artist Games VR Squash".\\{}
\textless{}/think\textgreater{}\\{}
\\{}
\textless{}tool\_call\textgreater{}\\{}
\{"name": "search", "arguments": \{"query": ["\textbackslash\{\}"Con Artist Games\textbackslash\{\}" \textbackslash\{\}"VR Squash\textbackslash\{\}""]\}\}\\{}
\textless{}/tool\_call\textgreater{}

\textless{}tool\_response\textgreater{}\\{}
A Google search for '"Con Artist Games" "VR Squash"' found 2 results:\\{}
\\{}
\#\# Web Results\\{}
1. [2017 - Year Stats - SteamSpy - SteamSpy](https://steamspy.com/year/2017)\\{}
\\{}
Con Artist Games, Con Artist Games. 1032, Just Fishing, Dec 15, 2017, N/A, N/A (N ... VR Squash 2017, Nov 20, 2017, \$18.99, N/A (N/A), 0 .. 20,000, 00:00 (00:00) ...\\{}
\\{}
2. [2017 - SteamSpy - All the data and stats about Steam games](http://2.steamspy.com/year/2017)\\{}
\\{}
Con Artist Games, Con Artist Games. 723, 80.08, Aug 15, 2017, N/A, N/A (N/A), 0 ... VR Squash 2017, Nov 20, 2017, \$18.99, N/A (N/A), 0 .. 20,000, 00:00 (00:00) ...\\{}
\textless{}/tool\_response\textgreater{}

\vspace{0.1em}
\begin{tcolorbox}[colback=gray!60, colframe=gray!60, sharp corners, boxrule=0pt, boxsep=2pt, left=0pt, right=0pt, top=2pt, bottom=2pt, halign=center, width=\linewidth]
\textbf{Step 93}
\end{tcolorbox}
\vspace{0.1em}

\textless{}think\textgreater{}\\{}
Thus SteamSpy appears to list VR Squash 2017 under Con Artist Games, which contradicts earlier info. Could be an error but for the purpose of this puzzle maybe it's correct.\\{}
\\{}
Thus answer: Development studio = Con Artist Games; VR title = VR Squash 2017.\\{}
\\{}
But we should verify VR Squash is indeed a VR game released in 2017. Already we have that.\\{}
\\{}
Thus final answer: The development studio is Con Artist Games, and their 2017 VR title is VR Squash (released November 20, 2017).\\{}
\\{}
Alternatively, they may refer to "VR Squash 2017". Provide that as title.\\{}
\\{}
Thus answer accordingly.\\{}
\textless{}/think\textgreater{}\\{}
\\{}
\textless{}answer\textgreater{}\\{}
The character described in your question—wearing a green hood—is from the Flash-based zombie survival game "The Last Stand," specifically appearing in both the original game and its sequel, "The Last Stand 2." This character is commonly referred to by fans as "The Last Stand Guy" or "LSG." The game is known for its pixel-art style and features a protagonist who dons a distinctive green hoodie while defending against waves of zombies in a post-apocalyptic setting.\\{}
\\{}
\#\# Identification of the Development Studio\\{}
\\{}
"The Last Stand" series was developed by the Australian indie studio **Con Artist Games**. Chris "Con" Condon, the creator of Con Artist Games, is responsible for both the initial releases and various sequels and remakes within the series. The company has been active since the mid-2000s, primarily producing browser-based and later PC games focused on action and strategy genres.\\{}
\\{}
\#\#\# Table: Key Games by Con Artist Games\\{}
\\{}
| Game Title                    | Initial Release Year | Genre                 |\\{}
|-------------------------------|---------------------|-----------------------|\\{}
| The Last Stand                | 2010                | Zombie Survival       |\\{}
| The Last Stand 2              | 2012                | Zombie Survival       |\\{}
| The Last Stand: Dead Zone     | 2014                | MMORPG/Action         |\\{}
| Warfare Online                | 2013                | Real-Time Strategy    |\\{}
| Warfare Legacy Collection     | 2025                | Remastered RTS        |\\{}
\\{}
\#\# Connection to VR Releases in 2017\\{}
\\{}
Your question also specifies that the studio developed a VR title in 2017. While "The Last Stand" and its other well-known works are not associated with virtual reality, Con Artist Games' broader activities during 2017 do include contributing to the VR gaming scene, albeit indirectly. According to comprehensive data from multiple databases and digital storefronts, the most notable VR title released in 2017 that fits the timeframe and platform is **VR Squash 2017**, which debuted on November 20, 2017, exclusively for HTC Vive.\\{}
\\{}
However, there appears to be some confusion in public records regarding the developer/publisher attribution. Most reputable sources attribute VR Squash 2017 to David Cseh, an independent developer, not to Con Artist Games. Nevertheless, industry aggregators such as SteamSpy have occasionally listed VR Squash 2017 alongside the portfolio of Con Artist Games, likely due to shared publishing or distribution channels, particularly on platforms like Armor Games Studios or Steam.\\{}
\\{}
Despite this ambiguity, the fact remains that:\\{}
\\{}
- Con Artist Games did **not** release a VR title directly under its own name in 2017.\\{}
- The popular VR title of that year, VR Squash 2017, was developed by David Cseh.\\{}
- There is no widely recognized connection between Con Artist Games and VR Squash 2017 outside of indirect references and misattributions on certain aggregator sites.\\{}
\\{}
If we strictly adhere to mainstream credits, then the studio best known for developing "The Last Stand" (and thus featuring the green-hooded character) did **not** directly publish a VR game in 2017. Instead, VR Squash 2017 stands as the most prominent VR launch for that year, but its creator is David Cseh, not Con Artist Games.\\{}
\\{}
\#\# Summary Table: Information Flow\\{}
\\{}
| Aspect                            | Details                                                                                       |\\{}
|------------------------------------|----------------------------------------------------------------------------------------------|\\{}
| Green Hooded Character             | Protagonist in "The Last Stand" and "The Last Stand 2"                                       |\\{}
| Game Developer                     | Con Artist Games                                                                             |\\{}
| Notable Non-VR Game                | The Last Stand (2010), The Last Stand 2 (2012), The Last Stand: Dead Zone (2014)             |\\{}
| Mainstream 2017 VR Title           | VR Squash 2017                                                                               |\\{}
| Reported VR Developer in 2017      | David Cseh                                                                                   |\\{}
| Direct VR Studio Link?             | Not officially documented; possibly misattributed by some databases                         |\\{}
\\{}
\#\# Conclusion\\{}
\\{}
- The character in the green hood is from **The Last Stand** and **The Last Stand 2**, developed by **Con Artist Games**.\\{}
- The studio **did not officially release a VR title in 2017** according to primary sources, although some digital aggregators may mistakenly associate them with **VR Squash 2017**.\\{}
- If considering all major VR releases in 2017, **VR Squash 2017** is the most prominent, but it was independently developed by **David Cseh** rather than Con Artist Games.\\{}
\\{}
In summary, the green-hooded character is a hallmark of Con Artist Games’ flagship titles, but the VR title from 2017 most frequently referenced does not belong to the same studio. For complete accuracy, the studio with the green-hooded protagonist is **Con Artist Games**, and there is no confirmed record of them releasing a VR title in 2017; if referencing VR Squash 2017, the credited developer is **David Cseh**.\textless{}/answer\textgreater{}

\end{tcolorbox}

%% file: Appendix/case_study_anchor.tex
\begin{tcolorbox}[
  title=Teacher Counterfactual Anchor Round Hypothesis,
  breakable
]

\begin{lstlisting}[
  basicstyle=\ttfamily\small,
  breaklines=true,
  columns=fullflexible,
  keepspaces=true
]
{
  "anchor_round": 4,
  "reason": "This is the critical strategic pivot where the agent incorrectly dismisses the results of the reverse image search from Round 3 and abandons the visual evidence path. The agent concludes the results are 'generic' and decides to switch to broad, ungrounded text searches ('green hood video game character'). This was a fatal methodological error. The reverse image search, while not perfect, provided specific (though incorrect) game names like 'EvoWorld.io' and 'Raid Heroes'. A better strategy would have been to investigate these leads or retry the visual search. Instead, the agent's switch to abstract text searching directly led it down a path of confirmation bias, where it latched onto the plausible-sounding but incorrect 'HawkStone' game in Round 11, causing the entire remainder of the trajectory to be a fruitless wild goose chase."
}
\end{lstlisting}

\end{tcolorbox}

%% file: Appendix/case_study_continual_generation.tex
\begin{tcolorbox}[
  title=Teacher Counterfactual Anchor Context Generation,
  breakable
]

\vspace{0.1em}
\begin{tcolorbox}[colback=gray!60, colframe=gray!60, sharp corners, boxrule=0pt, boxsep=2pt, left=0pt, right=0pt, top=2pt, bottom=2pt, halign=center, width=\linewidth]
\color{red}
\textbf{Step 4}
\color{black}
\end{tcolorbox}
\vspace{0.1em}

{\color{red}
<think>\\
The reverse image search on the cropped image returned several results. Let's analyze them:

1.  **Wizard Tiny Style 2D Character Sprites - CraftPix.net**: This looks like a stock asset pack, not a specific game. The character design is very similar, but it's likely a generic asset.

2.  **Pinterest - enemies for games...**: This also points to a collection of game assets, not a specific game.

3.  **YouTube - Before The Halloween Update All Animals Evolution With Patrick Reaper Ghostly Pumpkin (EvoWorld.io)**: This is a video about a game called `EvoWorld.io`. The character in the thumbnail looks very similar to the one in the query image.

}
{\color{red}

4.  **Fandom - Skins | EvoWorld.io Wiki**: This is a wiki page for `EvoWorld.io`, specifically about skins. This reinforces the idea that the character is from `EvoWorld.io`.

}
{\color{red}

\quad

}
{\color{red}

5.  **Fandom - Kaden | The Rewrittenverse Wiki**: This is a different wiki, and the character is slightly different, though similar.

6.  **YouTube - North-East Europe Is Kinda Toxic | Evoworld**: Another `EvoWorld.io` video.

7.  **YouTube - Grinding to Rank \#1 in West Europe 1 | EvoWorld Gameplay Live**: More `EvoWorld.io` content.

8.  **Fandom - Green Witcher | Raid Heroes Wiki**: This is for a game called `Raid: Shadow Legends`, and the character is named \"Green Witcher\". The art style is a bit different.

9.  **ResearchGate**: This seems irrelevant.

10. **Fandom - Cloaked Plant Wizard | Prodigy Game Wiki**: This is for a game called `Prodigy`. The character is similar but distinct.

From these results, the most likely game is **EvoWorld.io**. The character seems to be a skin or enemy in that game. Let's try to confirm the developer of `EvoWorld.io`. I will search for \"EvoWorld.io developer\".

</think>

\textless{}tool\_call\textgreater{}\\{}
\{"name": "search", "arguments": \{"query": ["\textbackslash\{\}"EvoWorld.io developer\textbackslash\{\}""]\}\}\\{}
\textless{}/tool\_call\textgreater{}
}
\end{tcolorbox}

%% file: Appendix/prompt_deep_research_agent.tex
\begin{tcolorbox}[
  title=System Prompt for Deep Research Agent,
  breakable
]
You are a deep research assistant. Your core function is to conduct thorough, multi-source investigations into any topic. You must handle both broad, open-domain inquiries and queries within specialized academic fields. For every request, synthesize information from credible, diverse sources to deliver a comprehensive, accurate, and objective response. When you have gathered sufficient information and are ready to provide the definitive response, you must enclose the entire final answer within \textcolor{black}{\textbf{<answer></answer>}} tags. \\

\# Tools
\\

You may call one or more functions to assist with the user query.

You are provided with function signatures within \textcolor{black}{\textbf{<tools></tools>}} XML tags:\\
\textcolor{black}{\textbf{<tools>}}\\

\{"type": "function", "function": \{"name": "\textcolor{purple}{\textbf{search}}", "description": "Perform Google web searches then returns a string of the top search results. Accepts multiple queries.", "parameters": \{"type": "object", "properties": \{"query": \{"type": "array", "items": \{"type": "string", "description": "The search query."\}, "minItems": 1, "description": "The list of search queries."\}\}, "required": ["query"]\}\}\}\\

\{"type": "function", "function": \{"name": "\textcolor{purple}{\textbf{visit}}", "description": "Visit webpage(s) and return the summary of the content.", "parameters": \{"type": "object", "properties": \{"url": \{"type": "array", "items": \{"type": "string"\}, "description": "The URL(s) of the webpage(s) to visit. Can be a single URL or an array of URLs."\}, "goal": \{"type": "string", "description": "The specific information goal for visiting webpage(s)."\}\}, "required": ["url", "goal"]\}\}\}\\

\{"type": "function", "function": \{"name": "\textcolor{purple}{\textbf{PythonInterpreter}}", "description": "Executes Python code in a sandboxed environment. To use this tool, you must follow this format:\\
1. The 'arguments' JSON object must be empty: \{\}.\\
2. The Python code to be executed must be placed immediately after the JSON block, enclosed within \textcolor{black}{\textbf{<code>}} and \textcolor{black}{\textbf{</code>}} tags.
\\\\
IMPORTANT: Any output you want to see MUST be printed to standard output using the print() function.
\\
\\
Example of a correct call:
<tool\_call>
\{"name": "PythonInterpreter", "arguments": \{\}\} \\
<code>
import numpy as np
\# Your code here
print(f"The result is: {np.mean([1,2,3])}")
</code>
</tool\_call>", "parameters": \{"type": "object", "properties": \{\}, "required": []\}\}\}\\

\{"type": "function", "function": \{"name": "\textcolor{purple}{\textbf{google\_scholar}}", "description": "Leverage Google Scholar to retrieve relevant information from academic publications. Accepts multiple queries. This tool will also return results from google search", "parameters": \{"type": "object", "properties": \{"query": \{"type": "array", "items": \{"type": "string", "description": "The search query."\}, "minItems": 1, "description": "The list of search queries for Google Scholar."\}\}, "required": ["query"]\}\}\}\\

\{"type": "function", "function": \{"name": "\textcolor{purple}{\textbf{parse\_file}}", "description": "This is a tool that can be used to parse multiple user uploaded local files such as PDF, DOCX, PPTX, TXT, CSV, XLSX, DOC, ZIP, MP4, MP3.", "parameters": \{"type": "object", "properties": \{"files": \{"type": "array", "items": \{"type": "string"\}, "description": "The file name of the user uploaded local files to be parsed."\}\}, "required": ["files"]\}\}\}\\

\{"type": "function", "function": \{"name": "\textcolor{purple}{\textbf{image2text}}", "description": "Analyzes images using an MLLM. Use this to obtain initial descriptions, extract visible text, or answer custom visual questions about image content. CRITICAL LIMITATION: MLLMs can only describe surface-level visual features. They cannot reliably deduce non-visual facts, exact identities, hidden attributes, or specific contextual background (e.g., technical specs, historical facts, lore, or exact origins) directly from appearance. STRATEGY: If the surface-level visual description is insufficient to answer the query (e.g., the question requires external knowledge or background information about an entity in the image), do not guess. Instead, use the 'bbox\_extractor' tool to locate the relevant visually distinct entity, and then use the 'crop\_and\_search' tool to perform a reverse image search to find the precise information needed.", "parameters": \{"type": "object", "properties": \{"image\_name": \{"type": "string", "description": "The URL or local file path of the image to be analyzed."\}, "query": \{"type": "string", "description": "Your custom task or question regarding the image content."\}\}, "required": ["image\_name", "query"]\}\}\} \\

\{"type": "function", "function": \{"name": "\textcolor{purple}{\textbf{bbox\_extractor}}", "description": "Given an image and a natural language query, output bounding boxes of relevant entities in normalized [0-1000] format. Note: You must only query for physical entities that are explicitly confirmed to exist in the image. Do not query for hypothetical or unseen components. CRITICAL STRATEGY: If the visual description provided by the image2text tool is insufficient to answer the query (e.g., the query requires external textual/factual information), or if the specific detail is too small or obscured, you MUST extract the bbox of the physical entity most relevant to answering the query.This provides a clear visual target for subsequent reverse image searches.", "parameters": \{"type": "object", "properties": \{"image\_name": \{"type": "string", "description": "Path to the local image"\}, "query": \{"type": "string", "description": "Instruction to detect the physical entity most relevant to answering the query, or its distinct parent entity if the specific detail is obscured."\}\}, "required": ["image\_name", "query"]\}\}\} \\

\{"type": "function", "function": \{"name": "\textcolor{purple}{\textbf{crop\_and\_search}}", "description": "Crops regions using bboxes and performs a visual reverse image search. PREREQUISITE: You MUST first call the bbox\_extractor tool to get the exact bbox coordinates of the entity before using this tool. Do not guess the coordinates. This is your MOST POWERFUL tool for identifying the background information of physical entities needed to answer the query when the image2text tool cannot provide a precise answer. By searching a cropped region of the relevant physical entity, you can discover web results and articles that contain the specific, granular textual information you need.", "parameters": \{"type": "object", "properties": \{"image\_name": \{"type": "string", "description": "Path to the local image"\}, "bbox": {"type": "array", "description": "Single bbox [x1,y1,x2,y2] or list of bboxes in normalized [0-1000] format"}\}, "required": ["image\_name", "bbox"]\}\}\}

\textcolor{black}{\textbf{</tools>}}

For each function call, return a json object with function name and arguments within \textcolor{black}{\textbf{<tool\_call></tool\_call>}} XML tags:
\textcolor{black}{\textbf{<tool\_call>}}
\{"name": <function-name>, "arguments": <args-json-object>\}
\textcolor{black}{\textbf{</tool\_call>}}
\\
Current date: 
\end{tcolorbox}

%% file: Appendix/prompt_counterfactual_hypothesis.tex
\begin{tcolorbox}[
  title=System Prompt for Teacher Model Criteria Counterfactual Anchor Round Hypothesis,
  breakable
]

\#\# Role

You are an expert at analyzing agent trajectories using counterfactual reasoning in tool-use environments.

\#\# Task

You will be given the full trajectory of an agent solving a task, where the final answer is incorrect. Your goal is to identify the \textbf{Counterfactual Anchor Round}.

\#\# What is a Counterfactual Anchor Round?

It is the specific round where a strategic pivot or a logical correction would have the maximum ``Butterfly Effect'' in steering the agent toward the correct solution.

\#\# Selection Criteria (Prioritized)

\begin{enumerate}
    \item \textbf{Downstream impact}: Prefer a round whose alternative local decision is likely to influence multiple subsequent steps in the trajectory.
    \item \textbf{Path redirection}: Prefer a round that acts as a branching point, where a different local decision may redirect the subsequent rollout toward a different reasoning or tool-use path.
    \item \textbf{Strategic leverage}: Prefer a round involving high-level agentic decisions, such as reasoning direction, tool selection, query formulation, evidence interpretation, information integration, or planning.
    \item \textbf{Counterfactual potential}: Prefer a round where replacing the original local decision under the same history appears most likely to improve the final task outcome.
\end{enumerate}

\#\# Instructions

\begin{enumerate}
    \item Sequential Audit: Examine each round’s \texttt{<think>}, \texttt{<tool\_call>}, and \texttt{<tool\_response>}.
    \item Causal Mapping: Consider how each step’s output influenced the next step’s reasoning.
    \item Avoid the ``First Mistake'' Trap: Do not just pick the earliest error if the agent later recovered or if that error was inconsequential.
    \item Avoid the ``Final Mistake'' Trap: Do not just pick the last step unless it is the only point of failure.
\end{enumerate}

\#\# Output Format

Please output your analysis \textbf{STRICTLY} as a valid JSON object. Do not include any introductory or concluding text outside of the JSON block.

\begin{lstlisting}[language=json]
{
  "anchor_round": <round_number_as_integer>,
  "reason": "Clearly explain why this round is the counterfactual anchor round. Specify whether the issue stems from incorrect reasoning, improper tool usage, misinterpretation of tool outputs, or missing/incomplete reasoning."
}
\end{lstlisting}

\end{tcolorbox}

%% file: Appendix/prompt_first_error_hypothesis.tex
\begin{tcolorbox}[
  title=System Prompt for Teacher Model First Error Counterfactual Anchor Round Hypothesis,
  breakable
]

\#\# Role

You are an expert at analyzing agent trajectories using counterfactual reasoning in tool-use environments.

\#\# Task

You will be given the full trajectory of an agent solving a task. The final answer is incorrect.
Your goal is to identify the \textbf{First Error Round}.

\#\# Definition

The First Error Round is the earliest assistant round where the agent makes a substantive mistake that contributes to the final incorrect answer.

\#\# Instructions

1. Audit the trajectory strictly in chronological order.

2. Find the earliest assistant round where the agent's reasoning, tool use, or interpretation first becomes wrong or unsupported.

\#\# Output Format

Please output your analysis \textbf{STRICTLY} as a valid JSON object. Do not include any introductory or concluding text outside of the JSON block.

\begin{lstlisting}[language=json]
{
  "anchor_round": <round_number_as_integer>,
  "reason": "Explain clearly why this is the first error round."
}
\end{lstlisting}

\end{tcolorbox}

%% file: Appendix/prompt_last_error_hypothesis.tex
\begin{tcolorbox}[
  title=System Prompt for Teacher Model Last Error Counterfactual Anchor Round Hypothesis,
  breakable
]

\#\# Role

You are an expert at analyzing agent trajectories using counterfactual reasoning in tool-use environments.

\#\# Task

You will be given the full trajectory of an agent solving a task. The final answer is incorrect.
Your goal is to identify the \textbf{Last Error Round}.

\#\# Definition

The Last Error Round is the latest assistant round where the agent makes a substantive mistake that contributes to the final incorrect answer.

\#\# Instructions

1. Audit the trajectory strictly in chronological order.

2. Find the latest assistant round where the agent's reasoning, tool use, interpretation, or final answer is wrong or unsupported.

\#\# Output Format

Please output your analysis \textbf{STRICTLY} as a valid JSON object. Do not include any introductory or concluding text outside of the JSON block.

\begin{lstlisting}[language=json]
{
  "anchor_round": <round_number_as_integer>,
  "reason": "Explain clearly why this is the last error round."
}
\end{lstlisting}

\end{tcolorbox}

%% file: Appendix/prompt_continual_generation.tex
\begin{tcolorbox}[
  title=System Prompt for Teacher Counterfactual Anchor Context Generation,
  breakable
]

\placeholderblock{ORIGINAL\_DEEP\_RESEARCH\_AGENT\_SYSTEM\_PROMPT}

Across all turns in this conversation, keep your output style uniform and self-consistent.

\end{tcolorbox}

%% file: Appendix/prompt_anchor_level_judge.tex
\begin{tcolorbox}[
  title=Prompt for Anchor-Level LLM-As-Judge Verification,
  breakable
]

You are an impartial judge evaluating whether a counterfactual anchor context was materially useful for helping a student agent solve a task.

[Question]
\{question\}

[Correct Answer]
\{correct\_answer\}

[Counterfactual Anchor Context]
\{anchor\_context\}

[Full Student Rollout After the Anchor Context]
\{rollout\_trajectory\}

Task: Determine whether the counterfactual anchor context made a valuable contribution to the student's final success.

Instructions:

1. Read the question, the correct answer, the counterfactual anchor context, and the full student rollout carefully.

2. Focus on whether the anchor context helped redirect the student's reasoning or tool-use trajectory toward the correct answer.

3. Answer “Yes” if the anchor context was materially useful for the final success. This includes cases where it provided useful evidence, corrected the reasoning direction, encouraged a better tool choice, improved query formulation, or opened a new information pathway that the student later used.

4. Answer “No” if the anchor context was irrelevant, unused, misleading, redundant, or if the student would likely have reached the correct answer without relying on it.

5. Reply with exactly one word: “Yes” or “No”. Do not include any explanation or additional text.

\end{tcolorbox}

%% file: Appendix/prompt_benchmark_evaluation.tex
\begin{tcolorbox}[
  title=Prompt for GAIA LLM-As-Judge Evaluation,
  breakable
]
You are an evaluation assistant. Please determine if the predicted answer is equivalent to the labeled answer.

\medskip
\noindent Question: \{question\}

\noindent Labeled Answer: \{correct\_answer\}

\noindent Predicted Answer: \{response\}

\medskip
\noindent Did the model give an answer \textbf{equivalent} to the labeled answer? Please respond with ``Correct'' if they are equivalent, or ``Incorrect'' if they are not equivalent. Do not include any other text.
\end{tcolorbox}

\begin{tcolorbox}[
  title=Prompt for BrowseComp LLM-As-Judge Evaluation,
  breakable
]
Judge whether the following [response] to [question] is correct or not based on the precise and unambiguous [correct\_answer] below.

\medskip
\noindent [question]: \{question\}

\noindent [response]: \{response\}

\medskip
\noindent Your judgement must be in the format and criteria specified below:

\medskip
\noindent \textbf{extracted\_final\_answer}: The final exact answer extracted from the [response]. Put the extracted answer as `None' if there is no exact, final answer to extract from the response.

\medskip
\noindent [correct\_answer]: \{correct\_answer\}

\medskip
\noindent \textbf{reasoning}: Explain why the extracted\_final\_answer is correct or incorrect based on [correct\_answer], focusing only on if there are meaningful differences between [correct\_answer] and the extracted\_final\_answer. Do not comment on any background to the problem, do not attempt to solve the problem, do not argue for any answer different than [correct\_answer], focus only on whether the answers match.

\medskip
\noindent \textbf{correct}: Answer `yes' if extracted\_final\_answer matches the [correct\_answer] given above, or is within a small margin of error for numerical problems. Answer `no' otherwise, i.e.\ if there is any inconsistency, ambiguity, non-equivalency, or if the extracted answer is incorrect.

\medskip
\noindent \textbf{confidence}: The extracted confidence score between 0\% and 100\% from [response]. Put 100 if there is no confidence score available.
\end{tcolorbox}

\begin{tcolorbox}[
  title=Prompt for VDR-Bench LLM-As-Judge Evaluation,
  breakable
]

You are an impartial judge evaluating whether a deep research report contains the correct answer.

[Question]
\{question\}

[Correct Answer]
\{correct\_answer\}

[Deep Research Report]
\{response]\}

Task: Determine if the deep research report contains the correct answer anywhere in its content.

Instructions:
1. Read through the entire research report carefully

2. Look for the correct answer anywhere in the report (it may be embedded in paragraphs, tables, or sections)

3. Check if the information in the report is consistent with the correct answer

4. The answer does NOT need to be in a specific format or labeled as "final answer"

5. Does the report contain the correct answer? Answer with “Correct” if the report contains the correct answer, “Incorrect” if it does not contain it or contradicts it. Do not include any other text

\end{tcolorbox}

%% file: Appendix/tool_description.tex
    \paragraph{Web search}

        A web search tool that executes Google searches and retrieves the top 10 results. It accepts natural language queries and returns structured data, including page titles, URLs and snippets.

    \paragraph{Web browse}

        A web browse tool (visit tool) that visits a specified URL and returns a summary. It accepts a target URL and a specific goal, uses Jina to convert the page content into Markdown, and then utilizes the DeepSeek-V3.2 summarization model to extract information relevant to the goal.

    \paragraph{Python interpreter}

        A Python interpreter tool that executes Python code within a secure sandbox.

    \paragraph{Scholar search}

        A scholar search tool that executes Google Scholar searches and retrieves the top 10 results. It accepts natural language queries and returns structured data, including page titles, URLs and snippets.

    \paragraph{File parser}

        A file parser tool that analyzes various files and extracts their content as plain text. It accepts both local files and URLs, converting them into a readable text format for further use.

    \paragraph{Image descriptor}

        An image description tool powered by the multimodal model Qwen2.5-72B-Instruct. It supports both local image files and image URLs, and answers user queries about images.

    \paragraph{Bbox extractor}

        A bounding box (bbox) extractor that uses the multimodal model Qwen2.5-72B-Instruct to locate specific entities within an image. It accepts both local image files and URLs, as well as entity queries, and returns the bounding box coordinates of the requested entity.

    \paragraph{Crop-and-image search}

        An image cropping and search tool that executes Google Lens and retrieves the top 10 results. It accepts an image as input and returns structured data, including titles, URLs, and snippets. The snippets are generated by the multimodal model Qwen2.5-72B-Instruct, which compares the similarities and differences between the query image and the webpage images, a design motivated by VC-STaR~\cite{pan2026through}.

    The Tongyi toolset includes a web search tool, a web browsing tool, a Python interpreter, a scholar search tool, and a file parser tool. The VDR toolset consists of a web search tool, a web browsing tool, a Python interpreter, and a visual search tool. Building on the Tongyi toolset, our toolset further incorporates an image description tool, a bounding-box extraction tool, and a crop-and-image-search tool.

%% file: Appendix/training_inference_details.tex
    \paragraph{Training hyperparameters}

\input{Tables/training_configuration}

        All experiments were conducted on a cluster of eight NVIDIA B200 (192GB) GPUs using the Slime framework~\cite{slime_github}. All training hyperparameters are listed in Table~\ref{tab:agentic_rl_hyperparameters}, and Table~\ref{tab:agentic_sft_hyperparameters}.

    \paragraph{Inference hyperparameters}

        As shown in Table~\ref{tab:inference_hyperparameters}, our hyperparameters are consistent with those of the Tongyi Deep Research Agent.

        \input{Tables/inference_configuration}

%% file: Tables/training_configuration.tex
\begin{table}[htbp!]
    \centering
    \caption{Key hyperparameters used in agentic RL training}
    \label{tab:agentic_rl_hyperparameters}
    \small
    \setlength{\tabcolsep}{6pt}
    \renewcommand{\arraystretch}{1.08}
    \begin{tabular}{ll ll}
    \toprule
    \multicolumn{4}{c}{\textbf{Agentic Rollout Configuration}} \\
    \midrule
    \textbf{Parameter} & \textbf{Value} & \textbf{Parameter} & \textbf{Value} \\
    \midrule
    Number of rollouts & 3,000 & Rollout batch size & 32 \\
    Samples per prompt & 16 & Global batch size & 512 \\
    Max response length & 10,000 & Max rollout turns & 100 \\
    Rollout temperature & 0.85 & Top-\(p\) & 0.95 \\
    Presence penalty & 0.0 &  &  \\
    \addlinespace[0.3em]
    
    \midrule
    \multicolumn{4}{c}{\textbf{Policy Optimization}} \\
    \midrule
    \textbf{Parameter} & \textbf{Value} & \textbf{Parameter} & \textbf{Value} \\
    \midrule
    RL algorithm & GRPO & Optimizer & Adam \\
    Learning rate & \(10^{-6}\) & LR schedule & Constant \\
    Weight decay & 0.1 & Adam \(\beta_1\) & 0.9 \\
    Adam \(\beta_2\) & 0.98 & KL loss coefficient & 0.00 \\
    Entropy coefficient & 0.00 & Clip ratio low & 0.20 \\
    Clip ratio high & 0.28 &  &  \\
    \addlinespace[0.3em]
    
    \midrule
    \multicolumn{4}{c}{\textbf{System and Parallelism}} \\
    \midrule
    \textbf{Parameter} & \textbf{Value} & \textbf{Parameter} & \textbf{Value} \\
    \midrule
    Actor GPUs & 8 & Rollout GPUs & 8 \\
    Tensor parallel size & 4 & Context parallel size & 2 \\
    Expert parallel size & 4 & Max tokens per GPU & 49,152 \\
    \bottomrule
    \end{tabular}
\end{table}

\begin{table}[htbp!]
    \centering
    \caption{Key hyperparameters used in agentic SFT training}
    \label{tab:agentic_sft_hyperparameters}
    \small
    \setlength{\tabcolsep}{6pt}
    \renewcommand{\arraystretch}{1.08}
    \begin{tabular}{ll ll}
    \toprule
    \multicolumn{4}{c}{\textbf{Supervised Training Configuration}} \\
    \midrule
    \textbf{Parameter} & \textbf{Value} & \textbf{Parameter} & \textbf{Value} \\
    \midrule
    Number of epochs & 1 & Global batch size & 32 \\
    \addlinespace[0.3em]

    \midrule
    \multicolumn{4}{c}{\textbf{Optimization Configuration}} \\
    \midrule
    \textbf{Parameter} & \textbf{Value} & \textbf{Parameter} & \textbf{Value} \\
    \midrule
    Optimizer & Adam & Learning rate & \(5 \times 10^{-7}\) \\
    LR schedule & Cosine & Minimum LR & \(1 \times 10^{-7}\) \\
    LR warmup fraction & 0.1 & Weight decay & 0.1 \\
    Adam \(\beta_1\) & 0.9 & Adam \(\beta_2\) & 0.95 \\
    \addlinespace[0.3em]

    \midrule
    \multicolumn{4}{c}{\textbf{System and Parallelism}} \\
    \midrule
    \textbf{Parameter} & \textbf{Value} & \textbf{Parameter} & \textbf{Value} \\
    \midrule
    Actor GPUs & 8 & Tensor parallel size & 4 \\
    Context parallel size & 2 & Expert parallel size & 4 \\
    Expert tensor parallel size & 1 & Max tokens per GPU & 49,152 \\
    \bottomrule
    \end{tabular}
\end{table}

%% file: Tables/inference_configuration.tex
\begin{table}[htbp!]
    \centering
    \caption{Key hyperparameters for agentic evaluation}
    \label{tab:inference_hyperparameters}
    \small
    \setlength{\tabcolsep}{6pt}
    \renewcommand{\arraystretch}{1.08}
    \begin{tabular}{ll}
    \toprule
    \textbf{Parameter} & \textbf{Value} \\
    \midrule
    Max response length & 10,000 \\
    Max rollout turns & 100 \\
    Temperature & 0.85 \\
    Top-\(p\) & 0.95 \\
    Presence penalty & 1.1 \\
    \bottomrule
    \end{tabular}
\end{table}